\documentclass[3p,a4paper,fleqn]{cas-dc}

\usepackage{multirow}
\usepackage{amsmath,amsfonts}
\usepackage{amssymb}
\usepackage{tikz}
\usepackage{algorithmic}
\usepackage{array}
\usepackage{soul}
\usepackage{xcolor}
\usepackage{algorithmic}
\usepackage{algorithm}
\usepackage{cancel}
\usepackage{hyperref}
\usepackage{stmaryrd}
\usepackage{textcomp}
\usepackage{stfloats}
\usepackage{url}
\usepackage{verbatim}
\usepackage{graphicx}
\usepackage{subfigure}
\usepackage{balance}
\usepackage{amsthm}
\usepackage{tikz}
\usetikzlibrary{matrix}
\usepackage{tabularx}
\usepackage{academicons}
\usepackage{orcidlink}
\usepackage{multirow}
\usepackage{tablefootnote}
\usepackage{chngcntr}
\usepackage{apptools}
\usepackage{soul,color}

\usepackage[many]{tcolorbox}
\tcbset{highlight style/.style={
  enhanced,
  breakable,
  boxrule=0mm,
  colback=green!30,
  sharp corners,
  boxsep=0pt,
  left=0pt, right=0pt, top=0pt, bottom=0pt,
  borderline west={1mm}{0pt}{green!30}
}}

\newtcolorbox{hlgreenbox}{highlight style}
\usepackage[authoryear]{natbib}

\newtheorem{theorem}{Theorem}
\newtheorem{lemma}[theorem]{Lemma}
\newtheorem{proposition}[theorem]{Proposition}
\newdefinition{rmk}{Remark}
\newproof{pf}{Proof}
\AtAppendix{\counterwithin{theorem}{section}}
\begin{document}
\let\WriteBookmarks\relax
\def\floatpagepagefraction{1}
\def\textpagefraction{.001}
\shorttitle{MonoKAN: Certified Monotonic Kolmogorov-Arnold Network}
\shortauthors{}

\title [mode = title]{MonoKAN: Certified Monotonic Kolmogorov-Arnold Network}

\author[1]{Alejandro Polo-Molina}[
                        orcid=0000-0001-7051-2288]
\cormark[1]
\ead{apolo@comillas.edu}


\affiliation[1]{organization={Institute for Research in Technology (IIT), ICAI School of Engineering, Universidad Pontificia Comillas},
                addressline={C/del Rey Francisco 4}, 
                city={Madrid},
                postcode={28008}, 
                state={Madrid},
                country={Spain}}

\author[1,2]{David Alfaya}[
                        orcid=0000-0002-4247-1498]
\ead{dalfaya@comillas.edu}

\author[1,3]{Jose Portela}[
                        orcid=0000-0002-7839-8982]
\ead{jportela@comillas.edu}
\credit{Writing – review \& editing, Validation, Supervision, Resources, Project administration, Methodology, Funding acquisition}
\affiliation[2]{organization={Department of Applied Mathematics, ICAI School of Engineering, Universidad Pontificia Comillas},
                addressline={C/Alberto Aguilera 25}, 
                city={Madrid},
                postcode={28015}, 
                state={Madrid},
                country={Spain}}

\affiliation[3]{organization={Department of Quantitative Methods, ICADE, Universidad Pontificia Comillas},
                addressline={C/Alberto Aguilera 23}, 
                city={Madrid},
                postcode={28015}, 
                state={Madrid},
                country={Spain}}

\cortext[cor1]{Corresponding author}
\maketitle

\begingroup\renewcommand\thefootnote{}\footnotetext{%
{\footnotesize © 2025 Alejandro Polo-Molina, David Alfaya, and Jose Portela.
This is the Accepted Manuscript of an article that will appear in \textit{Neural Networks}.
This version is made available under the terms of the Creative Commons
\textbf{Attribution-NonCommercial-NoDerivatives 4.0 International license (CC BY-NC-ND 4.0)}
\url{https://creativecommons.org/licenses/by-nc-nd/4.0/}.}%
}\addtocounter{footnote}{-1}\endgroup
\begin{abstract}
Artificial Neural Networks (ANNs) have significantly advanced various fields by effectively recognizing patterns and solving complex problems. Despite these advancements, their interpretability remains a critical challenge, especially in applications where transparency and accountability are essential. To address this, explainable AI (XAI) has made progress in demystifying ANNs, yet interpretability alone is often insufficient. In certain applications, model predictions must align with expert-imposed requirements, sometimes exemplified by partial monotonicity constraints. While monotonic approaches are found in the literature for traditional Multi-layer Perceptrons (MLPs), they still face difficulties in achieving both interpretability and certified partial monotonicity. Recently, the Kolmogorov-Arnold Network (KAN) architecture, based on learnable activation functions parametrized as splines,  has been proposed as a more interpretable alternative to MLPs. Building on this, we introduce a novel ANN architecture called MonoKAN, which is based on the KAN architecture and achieves certified partial monotonicity while enhancing interpretability. To achieve this, we employ cubic Hermite splines, which guarantee monotonicity through a set of straightforward conditions. Additionally, by using positive weights in the linear combinations of these splines, we ensure that the network preserves the monotonic relationships between input and output. Our experiments demonstrate that MonoKAN not only enhances interpretability but also improves predictive performance across the majority of benchmarks, outperforming state-of-the-art monotonic MLP approaches.
\end{abstract}


\section{Introduction}
Artificial neural networks (ANNs) are the backbone of modern artificial intelligence \citep{Lecun2015DeepLearning,Goodfellow2016DeepLearning}. These computational systems are designed to recognize patterns and solve complex problems through learning from data, making them highly effective for tasks such as image and speech recognition \citep{Hinton2012DeepGroups}, predictive analytics \citep{Liu2017AApplications} or many others \citep{Sarvamangala2022ConvolutionalSurvey,Xu2020ApplicationIntelligence}. By mimicking the brain's ability to process information and adapt through experience, ANNs have revolutionized fields ranging from computer vision to autonomous systems \citep{Sarvamangala2022ConvolutionalSurvey,Voulodimos2018DeepReview}, and their development continues to drive forward the capabilities of machine learning and artificial intelligence as a whole.

Despite their impressive capabilities, ANNs face significant challenges regarding interpretability. As ANNs grow more complex, their decision-making processes become increasingly opaque, often described as "black boxes" due to the difficulty in understanding how specific inputs are translated into outputs. This lack of transparency can be problematic in critical applications such as healthcare or finance, where understanding the rationale behind decisions is crucial for trust and accountability \citep{Cohen2021Black-BoxFinance,Tjoa2021AXAI}. Furthermore, the complexity of ANNs makes it hard to find and fix biases in the models, which can lead to unfair or harmful results. Addressing these interpretability issues is essential to ensure that ANNs can be safely and effectively integrated into high-stakes environments.

In response to the interpretability challenges of ANNs, the field of explainable artificial intelligence (XAI) has grown substantially in the last decades. Numerous studies have emerged aiming to demystify their inner workings \citep{Zhang2020AInterpretability,Pizarroso2022NeuralSens:Networks,Morala2023NN2Poly:Networks}.
These studies represent critical strides toward making neural networks more transparent and trustworthy, facilitating their adoption in fields where understanding and accountability are paramount.

However, it is often the case where interpretability alone is insufficient in some critical applications \citep{Rudin2019StopInstead}. Therefore, in some fields, it is a requisite to certify that the model predictions align with some requirements imposed by human experts \citep{Cohen2021Black-BoxFinance}. Partial monotonicity is an example where incorporating prior knowledge from human experts into the model might sometimes be necessary. For instance, in university admissions, it is reasonable to expect that, all other variables being equal, an applicant with a higher GPA should have a higher probability of being accepted. If the model's predictions do not follow this monotonic relationship, it could lead to unfair and unethical admission decisions. For instance, an applicant with a 4.0 GPA being rejected while an applicant with a 3.0 GPA is accepted, all other factors being equal, would be seen as unfair and could indicate bias in the model. 

Partial monotonicity also plays a key role in domains such as healthcare, finance or criminal justice \citep{Wang2020DeontologicalConstraints}. In organ transplant allocation, for example, policies often prioritize sicker patients with higher urgency scores to receive transplants sooner \citep{Iserson2007TriageTypes}, as is the case with the United States Transplant Board \citep{Bernstein2017WhoPost}. A model violating this principle, by assigning lower transplant priority to a patient with higher medical need, could result in unethical outcomes. Besides, in criminal sentencing, many legal systems impose stricter penalties on repeat offenders compared to first-time offenders \citep{Allen2016CorrectionsIntroduction}. A model used for risk assessment that fails to respect this monotonic relationship could produce recommendations that are legally and socially unacceptable. 
Consequently, certified monotonic ML models allow users to enforce such domain-specific monotonic constraints during training, helping ensure that the model adheres to expected input-output relations.

As a result, the training of partial monotonic ANNs has become a prominent area of research in recent years. To tackle this issue, two primary strategies have emerged \citep{Liu2020CertifiedNetworks}. First of all, there are some studies that enforce monotonicity by means of a regularization term that guides the ANNs training towards a partial monotonic solution \citep{Sivaraman2020Counterexample-GuidedNetworks,Gupta2019HowFlexibility,Monteiro2022MonotonicityClassification}. However, these approaches verify monotonicity only on a finite set of points, and hence none of the previous studies can certify the enforcement of partial monotonic constraints across all possible input values. Therefore, it is necessary to use some external certification algorithm after the training process to guarantee partial monotonicity. Regarding this type of algorithm, few examples are found in the literature \citep{Liu2020CertifiedNetworks, Polo-Molina2025AAI}. On the other hand, some studies propose designing constrained architectures that inherently ensure monotonicity \citep{Runje2023ConstrainedNetworks,Daniels2010MonotoneNetworks,You2017DeepFunctions,Nolte2022ExpressiveNetworks}. Although these methods can guarantee partial monotonicity, they often come with the trade-off of being overly restrictive or complex and challenging to implement \citep{Liu2020CertifiedNetworks}. 

Even though some of the aforementioned methods can lead up to certified partial monotonic ANNs, traditional Multi-layer Perceptron (MLP) architectures still have significant difficulties with interpretability. The complex and often opaque nature of the connections and weight adjustments in MLPs makes it challenging to understand and predict how inputs are being transformed into outputs. Therefore, existing approaches to obtaining monotonic MLPs hardly generate both interpretable and certified partial monotonic ANNs, often requiring post-hoc interpretability methods. 

To address some of the aforementioned difficulties related to interpretability, a new ANN architecture, called the Kolmogorov-Arnold Network (KAN), has been recently proposed \citep{Liu2024KAN:Networks}. Unlike traditional MLPs, which rely on the universal approximation theorem, KANs leverage the Kolmogorov-Arnold representation theorem. This theorem states that any multivariate continuous function can be decomposed into a finite combination of univariate functions, enhancing the interpretability of the network.

However, the functions depicted by the Kolmogorov-Arnold theorem can be non-smooth, even fractal, and may not be learnable in practice \citep{Liu2024KAN:Networks}. Consequently, a KAN with the width and depth proposed by the Kolmogorov-Arnold theorem is often too simplistic in practice to approximate any function arbitrarily well using smooth splines. 

Therefore, although the use of the Kolmogorov-Arnold representation theorem for ANNs was already studied \citep{Sprecher2002Space-fillingNetworks,Koppen2002OnNetwork}, the major breakthrough occurred when \citet{Liu2024KAN:Networks} established the analogy between MLPs and KANs. In MLPs, the notion of a layer is clearly defined, and the model's power comes from stacking multiple layers to form deeper architectures. Similarly, defining a KAN layer allows for the creation of deep KANs through layer stacking, which significantly enhances the model’s ability to capture increasingly complex functions.

Schematically, each KAN layer is composed of a set of nodes, with each node connected to all preceding nodes via activation functions on the edges. These univariate activation functions are the components subjected to training. Then, the outputs of these activation functions are aggregated to determine the node's output. According to \citep{Liu2024KAN:Networks}, this approach not only enhances interpretability by allowing visualization of relationships between variables, but also demonstrates faster neural scaling laws compared to MLPs due to its ability to decompose complex functions into simpler ones. Additionally, it can improve performance on numerous problems compared to MLPs \citep{Poeta2024AData, Xu2024Kolmogorov-ArnoldInterpretability}.

Building on these advantages, this paper proposes a novel KAN architecture called MonoKAN that forces the resulting KAN to be certified partially monotonic across the entire input space. To do so, while the original formulation of KANs proposes the use of B-splines, this paper replaces them with cubic Hermite splines and imposes constraints on their coefficients to ensure monotonicity. This substitution allows for more flexible and general imposition of monotonic conditions. An intuitive rationale for this change is that, while it is possible to achieve monotonicity with a combination of B-splines within a specific interval, B-splines are not inherently monotonic. In contrast, cubic Hermite splines can be imposed to be monotonic naturally \citep{Fritsch1980MonotoneInterpolation, Arandiga2022MonotoneGradient}, making them a more appropriate choice for ensuring the desired monotonic properties in the MonoKAN architecture.

Therefore, MonoKAN enhances the capability of the KAN framework by leveraging the intrinsic properties of cubic Hermite splines to achieve certified partial monotonicity. Consequently, the proposed MonoKAN architecture is able to encompass both the enhanced interpretability that the KAN architecture presents with certified partial monotonicity. To the authors’ knowledge, this is the first time that a monotonic approach for a KAN has been proposed.

The paper is structured as follows: Section \ref{sec:kan_networks} presents the KAN methodology that will be later used in Section \ref{sec:monokan} as the base to generate certified partial monotonic KANs. Besides, in Section \ref{sec:experiments}, the experiments and corresponding results are detailed, demonstrating that the proposed approach surpasses the state-of-the-art methods in the majority of the experiments. Moreover, in Section \ref{sec:computational_implementation_details}, the computational complexity of the proposed method and the comparison with the state-of-the-art models is presented. Lastly, the main contributions are summarized in Section \ref{sec:conclusion}. Moreover, the code of the proposed algorithm and the results are available at \href{https://github.com/alejandropolo/MonoKAN}{\url{https://github.com/alejandropolo/MonoKAN}}
\vspace{-0.4cm} 
\section{Kolmogorov-Arnold Networks}\label{sec:kan_networks}
As mentioned before, while Multi-Layer Perceptrons (MLPs) draw their inspiration from the universal approximation theorem, our attention shifts to the Kolmogorov-Arnold representation theorem. 

The Kolmogorov-Arnold representation theorem states that any multivariate continuous function can be expressed as a finite sum of continuous functions of a single variable. Specifically, for any function $f: [0,1]^n \rightarrow \mathbb{R}$, there exist univariate functions $\phi_{q,p}:[0,1] \rightarrow \mathbb{R}$ and $\Phi_q: \mathbb{R} \rightarrow \mathbb{R}$ such that 
\begin{equation}\label{eq:kan_teo}
f(x_1, x_2, \ldots, x_n) = \sum_{q=1}^{2n+1} \Phi_q \left( \sum_{p=1}^n \phi_{q,p}(x_p) \right).
\end{equation}

The major breakthrough presented in \citep{Liu2024KAN:Networks} comes from recognizing the similarities between MLPs and KAN. Just as MLPs increase their depth and expressiveness by stacking multiple layers, KANs can similarly enhance their predictive power through layer stacking once a KAN layer is properly defined.

To understand this concept further, a KAN layer with $n_{\rm in}$ inputs and $n_{\rm out}$ outputs can be described as a matrix of 1D functions
\begin{align*}
    {\mathbf\Phi}=\{\phi_{q,p}\},\quad p=1,2,\cdots,n_{\rm in},\quad q=1,2\cdots,n_{\rm out},
\end{align*}
where $\phi_{q,p}(\cdot)$ are functions parameterized by learnable coefficients.
Consequently, applying a KAN layer with $n_{\rm in}$ inputs and $n_{\rm out}$ outputs to an input $\mathbf{x} \in \mathbb{R}^{n_{\rm in}}$ is defined through the following action of the matrix of functions. 
\begin{align*}
    {\mathbf\Phi} (\mathbf{x}) &= \left( \phi_{q,p} \right)_{1\leq p\leq n_{\rm in}, 1\leq q\leq n_{\rm out}} \cdot (x_p)_{1\leq p\leq n_{\rm in}}\\  &=  \sum_{p=1}^{n_{\rm in}} \phi_{q,p} (x_p),\quad q=1,2,\cdots,n_{\rm out}.
\end{align*}
Accordingly, the Kolmogorov-Arnold theorem (Eq. \eqref{eq:kan_teo}) can be represented within the KAN framework as a composition of a KAN layer with $n_{\rm in}=n$ and $n_{\rm out}=2n+1$ and a KAN layer with $n_{\rm in}=2n+1$ and $n_{\rm out}=1$. 

Given that all functions to be learned are univariate, we can approximate each 1D function as a spline curve with learnable coefficients. However, it is important to note that the functions $\phi_{q,p}(\cdot)$ specified by the Kolmogorov-Arnold theorem are arbitrary. In practice, though, a specific class of functions parameterized by a finite number of parameters is typically used. This practical consideration justifies the need of using more KAN layers than just the proposed by Eq. \eqref{eq:kan_teo}.

Adopting the notation from \citep{Liu2024KAN:Networks}, we define the structure of a KAN as $[n_0,n_1,\cdots,n_L]$, where $n_l$ denotes the number of nodes in the $l^{\rm th}$ layer. The $i^{\rm th}$ neuron in the $l^{\rm th}$ layer is represented by $(l,i)$, and its activation value by $x_{l,i}$. Between layer $l$ and layer $l+1$, there are $n_ln_{l+1}$ activation functions and the activation function connecting $(l,i)$ and $(l+1,j)$ is denoted by
\begin{align*}
    &\phi_{l,j,i}(\cdot),\quad l=0,\cdots, L-1,\\&\quad i=1,\cdots,n_{l},\quad j=1,\cdots,n_{l+1}.
\end{align*}
The pre-activation input for $\phi_{l,j,i}(\cdot)$ is $x_{l,i}$, while its post-activation output is represented by $\tilde{x}_{l,j,i}:= \phi_{l,j,i}(x_{l,i})$. Moreover, the activation value of the neuron $(l+1,j)$ is then calculated as the sum of all incoming post-activation values:
\begin{equation}\label{eq:post_activation}
    x_{l+1,j} =  \sum_{i=1}^{n_l} \tilde{x}_{l,j,i} = \sum_{i=1}^{n_l}\phi_{l,j,i}(x_{l,i}), \qquad j=1,\cdots,n_{l+1}.
\end{equation}
Therefore, the KAN layer can be stated in its matrix form as
\begin{equation}\label{eq:kanforwardmatrix}
    \mathbf{x}_{l+1} = 
    \underbrace{\begin{pmatrix}
        \phi_{l,1,1}(\cdot) & \phi_{l,1,2}(\cdot) & \cdots & \phi_{l,1,n_{l}}(\cdot) \\
        \phi_{l,2,1}(\cdot) & \phi_{l,2,2}(\cdot) & \cdots & \phi_{l,2,n_{l}}(\cdot) \\
        \vdots & \vdots & & \vdots \\
        \phi_{l,n_{l+1},1}(\cdot) & \phi_{l,n_{l+1},2}(\cdot) & \cdots & \phi_{l,n_{l+1},n_{l}}(\cdot) \\
    \end{pmatrix}}_{\mathbf{\Phi}_l}
    \mathbf{x}_{l},
\end{equation}
where ${\mathbf \Phi}_l$ is the function matrix corresponding to the $l^{\rm th}$ KAN layer. Consequently, a KAN with $L$-layers can be described as a composition of the function matrices ${\mathbf \Phi}_l, 0 \leq l < L$, such that for a given input vector \(\mathbf{x} \in \mathbb{R}^{n_0}\), the output of the KAN is: 
\begin{equation*}
    {\rm KAN}(\mathbf{x}) = \left(\mathbf{\Phi}_{L-1}\circ\cdots\circ\mathbf{\Phi}_{1}\circ \mathbf{\Phi}_{0}\right)(\mathbf{x}).
\end{equation*}

\section{MonoKAN}\label{sec:monokan}

This section introduces the necessary theoretical development and the proposed algorithm for generating a set of sufficient conditions to ensure that a KAN is partially monotonic w.r.t. a specific subset of input variables. First of all, the concept of partial monotonicity will be explained, as well as the way to ensure monotonicity in cubic Hermite splines. Subsequently, the main theorem outlining the sufficient conditions for a KAN to be partially monotonic will be presented. Finally, the proposed algorithm to ensure that a KAN meets these conditions will be described.

\subsection{Partial Monotonicity}
To begin with, let us start by presenting the concept of partial monotonicity. Intuitively, a function $f: \mathbb{R}^n \rightarrow \mathbb{R}$ is increasing (resp. decreasing) partially monotonic w.r.t the $r^{th}$ input, with $1 \leq r \leq n$,  whenever the output increases (decreases) if the $r^{th}$ input increases. Mathematically speaking, a function $ f:\mathbb{R}^n \rightarrow \mathbb{R} $ is increasing (resp. decreasing) partially monotonic w.r.t. its $r^{th}$ input if
\begin{equation}\label{eq:mono}
          f(x_1,\dots,x_{r},\dots,x_n) \leq f(x_1,\dots,x'_{r}\dots,x_n) \, , \forall \ x_r \leq x'_r\,
\end{equation}
\begin{equation*}
          (\text{resp.} \; f(x_1,\dots,x_{r},\dots,x_n) \geq f(x_1,\dots,x'_{r}\dots,x_n)).
\end{equation*}
Therefore, $f$ will be partially monotonic w.r.t. a subset of its input variables $\{x_{i_1},\dots,x_{i_k}\}$ where $0 \leq k \leq n $, if Eq. \eqref{eq:mono} holds for each $ i_j $ simultaneously with $ j \in \{1,\dots,k\} $.

\subsection{Monotonic Cubic Hermite Splines}
As mentioned before, each function to be learned in a KAN layer is univariate, allowing for various parameterization methods for each 1D function. While the original formulation presented in \citep{Liu2024KAN:Networks} employs B-splines to approximate these univariate functions, in this paper, it is proposed the use of cubic Hermite splines. The advantage of cubic Hermite splines lies in the well-known sufficient conditions for monotonicity \citep{Fritsch1980MonotoneInterpolation, Arandiga2022MonotoneGradient}. Besides, for a sufficiently smooth function and  a fine enough grid, the resulting cubic Hermite spline converges uniformly to the desired function \citep{Hall1976OptimalInterpolation}. 

A cubic Hermite spline, or cubic Hermite interpolator, is a type of spline where each segment is a third-degree polynomial defined by its values and first derivatives at the endpoints of the interval it spans. Consequently, the spline is $C^1$ continuous within the interval of definition.

To formally define a cubic Hermite spline, consider a set of knots $x_k$, values $y_k$ and derivative values $m_k$ at each of the knots $x_k$ given by  $\mathcal{X}= \{(x_k, y_k, m_k) \mid \forall \; k \in I = \{1,2,\dots, n\} \}$. Then, the cubic Hermite spline $p$ is a set of $n-1$ cubic polynomials such that, in each subinterval $I_k = [x_k,x_{k+1}]$, it is verified that
\begin{align}\label{eq:spline_def_condition}
p(x_k) &= y_k, \forall \; k \in I \\
p'(x_k) &= m_k, \forall \; k \in I \nonumber.
\end{align}

\begin{figure}[t]
    \centering
    \includegraphics[width=0.7\linewidth]{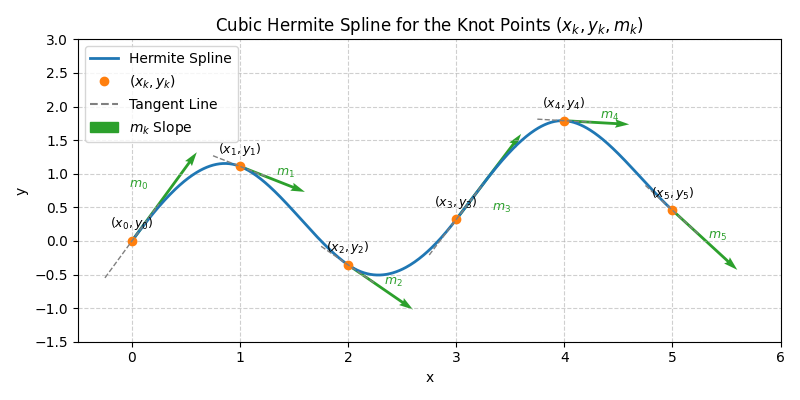}
    \caption{Illustration of a cubic Hermite spline. The spline smoothly interpolates the function values while respecting the slope (derivative) constraints at each knot.}
    \label{fig:hermite_spline}
\end{figure}

Therefore, the above conditions ensure that the spline matches both the function values and the slopes at each data point. An intuitive example illustrating this behavior is shown in Figure \ref{fig:hermite_spline}, where the spline passes through the knots and aligns with the expected derivatives.

Furthermore, on each subinterval $I_k=[x_k, x_{k+1}]$, the cubic Hermite spline $p$ can be expressed in its Hermite form as
\begin{multline*}
    p(t) = h_{00}(t)y_k + h_{10}(t)(x_{k+1} - x_k)m_k + \\ 
    h_{01}(t)y_{k+1} + h_{11}(t)(x_{k+1} - x_k)m_{k+1},
\end{multline*}
where $t = \frac{x - x_k}{x_{k+1} - x_k}$ 
and $h_{00}, h_{10}, h_{01}, h_{11}$ are the Hermite basis functions defined as follows
\begin{align*}
h_{00}(t) &= 2t^3 - 3t^2 + 1, \\
h_{10}(t) &= t^3 - 2t^2 + t, \\
h_{01}(t) &= -2t^3 + 3t^2, \\
h_{11}(t) &= t^3 - t^2.
\end{align*}
Once the terminology of cubic Hermite splines has been established, we now consider the conditions required for the resulting spline to be monotonic as shown in \citet{Fritsch1980MonotoneInterpolation}. According to Eq. \eqref{eq:spline_def_condition}, to achieve an increasing (resp. decreasing) monotonic cubic Hermite spline, it is necessary that $y_k \leq y_{k+1}, \; \forall \; k \in I \; (\text{resp. } y_k \geq y_{k+1}, \; \forall \; i \in I) $. Additionally, to ensure monotonicity, it is clear that considering 
\begin{equation*}
d_k = \frac{y_{k+1} - y_k}{x_{k+1} - x_k},
\end{equation*}
the slope of the secant line between two successive points $x_k$ and $x_{k+1}$, then the derivative at each point within the interval $I_k$ must match the sign of $d_k$ to maintain monotonicity. Specifically, if $d_k = 0$, then both $m_k$ and $m_{k+1}$ must also be zero, as any other configuration would disrupt monotonicity between $x_k$ and $x_{k+1}$. These conditions, stated in the following lemma, establish necessary conditions for a cubic Hermite spline to be monotonic.

\begin{lemma}[Necessary conditions for monotonicity, {{\cite[Theorem~1.1]{Arandiga2022MonotoneGradient}}}]
\label{teo:nec_condition_hermite}
    Let $p_k$ be an increasing (resp. decreasing) monotone cubic Hermite spline of the data $\mathcal{X} = \{(x_k, y_k, m_k), (x_{k+1}, y_{k+1}, m_{k+1})\}$ such that the control points verify that $y_k \leq y_{k+1}$ (resp. $y_k \geq y_{k+1}$). Then
\begin{align*}
m_k &\geq 0 \quad \text{and} \quad m_{k+1} \geq 0. \\
\big(\text{resp. } m_k &\leq 0 \quad \text{and} \quad m_{k+1} \leq 0\big)
\end{align*}
Moreover, if $d_k = 0$ then $p_k$ is monotone (in fact, constant) if and only if $m_k = m_{k+1} = 0$.
\end{lemma}

For the more general case when $d_k \neq 0$, \citet{Fritsch1980MonotoneInterpolation} introduced the parameters $\alpha_k$ and $\beta_k$, defined as
\begin{align*}
\alpha_k &:= \frac{m_k}{d_k}, \\
\beta_k &:= \frac{m_{k+1}}{d_k}.
\end{align*}

These parameters provide the necessary framework for establishing sufficient conditions for monotonicity. 
\begin{lemma}[Sufficient conditions for monotonicity, {{\cite[Lemma 2 and \S 4]{Fritsch1980MonotoneInterpolation}}}]\label{teo:suf_condition_hermite}
Let $I_k = [x_k, x_{k+1}]$ be an interval between two knot points and $p_k$ be a cubic Hermite spline of the data $\mathcal{X} = \{(x_k, y_k, m_k), (x_{k+1}, y_{k+1}, m_{k+1})\}$ such that the control points verify that $y_k < y_{k+1}$ (resp. $y_k > y_{k+1}$). Then, the cubic Hermite spline $p_k$ is increasingly (resp. decreasingly) monotone on $I_k$ if
\begin{align*}
\alpha_k &:= \frac{m_k}{d_k} \geq 0, & \beta_k &:= \frac{m_{k+1}}{d_k} \geq 0 \\
\big(\text{resp. } \alpha_k &:= \frac{m_k}{d_k} \leq 0, & \beta_k &:= \frac{m_{k+1}}{d_k} \leq 0\big) \\
\end{align*}
and
\begin{equation*}
\alpha_k^2 + \beta_k^2 \leq 9.
\end{equation*}
\end{lemma}

By adhering to these conditions, one can ensure that the cubic Hermite spline remains monotonic over its entire domain.

\subsection{Mathematical Certification of Partial Monotonicity of a KAN}\label{sec:math_certification}

Having introduced the definition of partial monotonicity and the necessary conditions for a cubic Hermite spline to be monotonic, we now present the main theoretical result, which provides a set of sufficient conditions for a KAN to be certified partial monotonic. For simplicity, we will assume that the KAN is partially monotonic with respect to the $r^{th}$ input. Therefore, when handling multiple monotonic features, the same conditions applied to the $r^{th}$ input will be applied to each monotonic feature.

Recall that, according to Eq. \eqref{eq:kanforwardmatrix}, a KAN can be described as a combination of univariate functions, parametrized in this paper as cubic Hermite splines, followed by a multivariate sum. Since a linear combination of monotonic functions with positive coefficients is monotonic, and the composition of monotonic functions is also monotonic (see Proposition \ref{ap:prop_compo_mono}), we propose constructing a partially monotonic KAN by ensuring that each of the cubic Hermite splines in the KAN is monotonic.

\begin{figure}[!t]
\centering
\includegraphics[width=\linewidth]{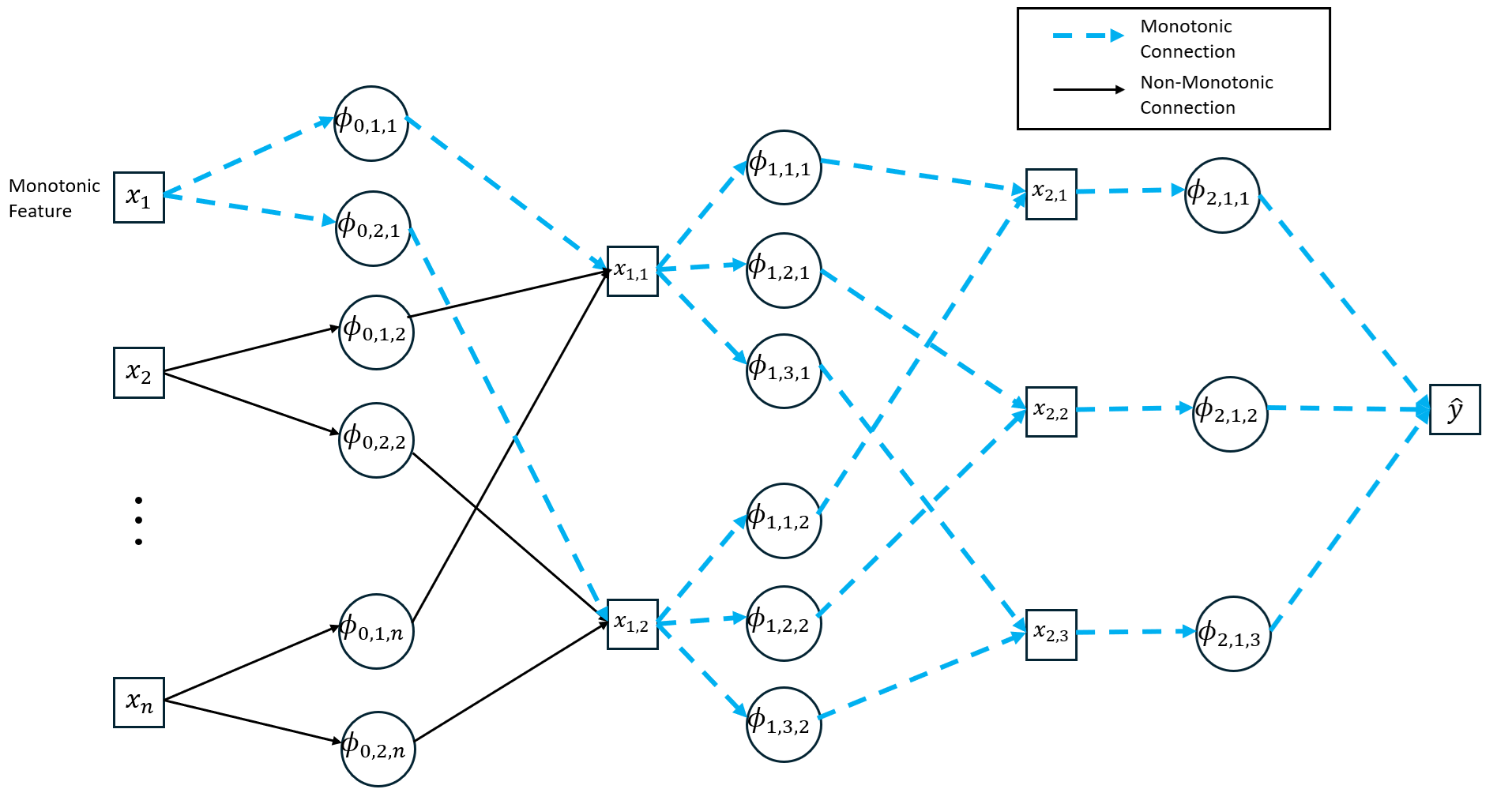}
\caption{Scheme of monotonic and non-monotonic connections of a partial monotonic KAN w.r.t the first input with layers $[n,2,3,1]$}
\label{fig:monokan_scheme}
\end{figure}

To achieve this, consider a KAN with $ n_0 = n $ inputs, expected to be increasingly (decreasingly) partially monotonic with respect to the $ r^{th} $ input, where $ 1 \leq r \leq n $. Consequently, to obtain an increasingly (decreasingly) partially monotonic KAN with respect to the $ r^{th} $ input, it is sufficient to ensure that the $ n_1 $ activations originating from the $ r^{th} $ input are increasingly (decreasingly) monotonic and that any of the following neurons, where the output of the activation functions generated by the $ r^{th} $ input is considered as part of the input, must also be increasingly monotonic. This idea of this procedure is illustrated in Figure \ref{fig:monokan_scheme}, which provides an example of a partial monotonic KAN.

\begin{figure}[t]
    \centering
    \includegraphics[width=0.9\linewidth]{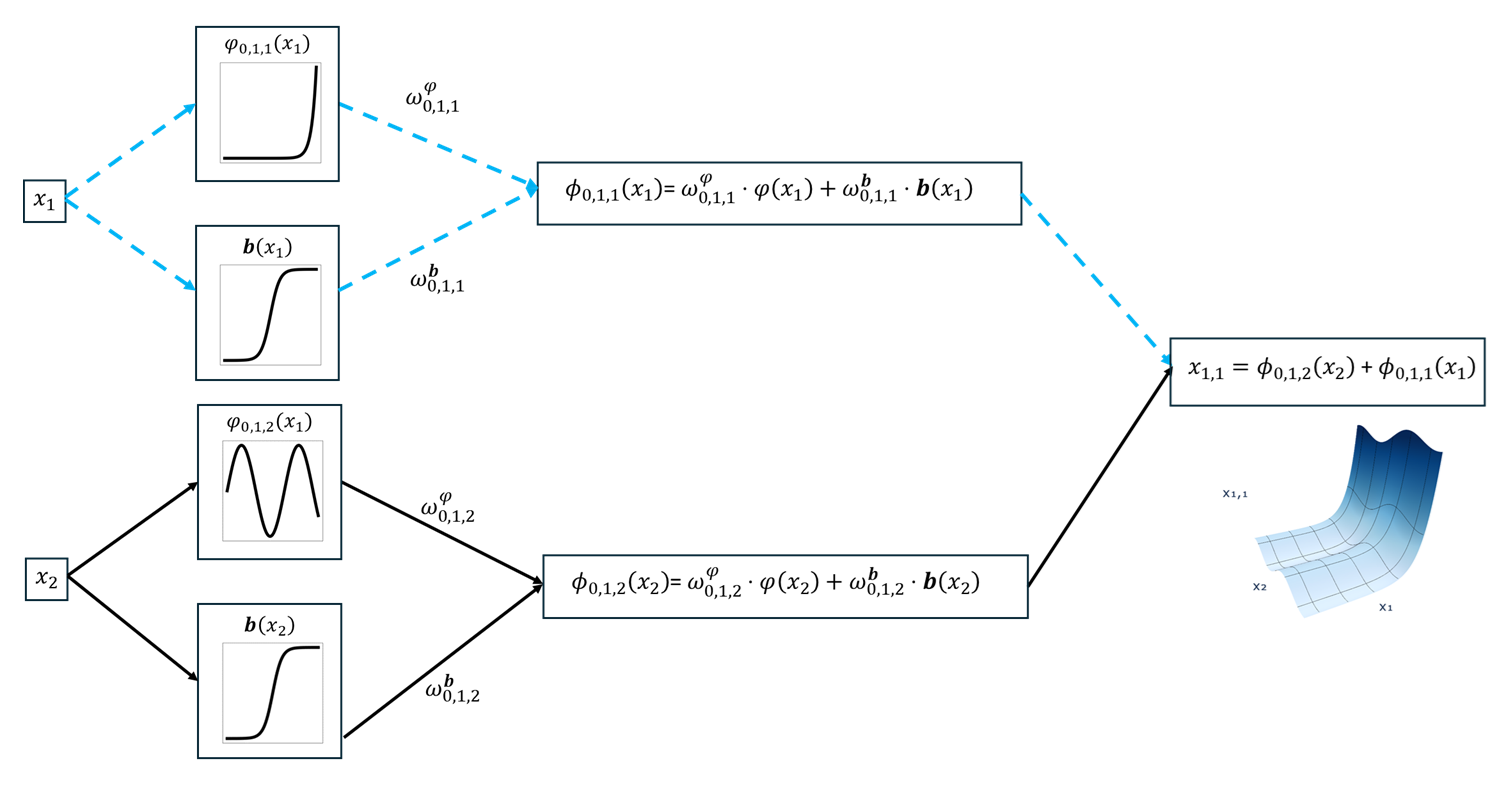}
    \caption{Illustration of a simplified MonoKAN, partially monotonic w.r.t. the first input, with two inputs and one output. Each connection between neuron $(l,i)$ and $(l+1,j)$ includes a cubic Hermite spline $\varphi_{l,j,i}$ and a base activation $\mathbf{b}$, weighted respectively by $\omega^\varphi_{l,j,i}$ and $\omega^\mathbf{b}_{l,j,i}$. The outputs of these functions are summed to produce the final activation.}
    \label{fig:example_simple_KAN}
\end{figure}

On the other hand, as mentioned in \citep{Liu2024KAN:Networks}, the activation value of the $(l+1,j)$ node is not computed merely as a sum of spline outputs of the previous layer, but rather as a combination of spline transformations and a standard activation function. Specifically, each activation $\phi_{l,j,i}$ is defined as a linear combination of a spline $\varphi_{l,j,i}$ and a base activation function $\mathbf{b}$ (e.g., Sigmoid or SiLU), both evaluated at the pre-activation $x_{l,i}$. This construction is visually detailed in Figure~\ref{fig:example_simple_KAN}, where the roles of $\varphi_{l,j,i}$ and $\mathbf{b}$, as well as their respective weights $\omega^\varphi_{l,j,i}$ and $\omega^b_{l,j,i}$, are explicitly depicted. Therefore, Eq.~\eqref{eq:post_activation} is transformed into:
\begin{equation}
\begin{aligned}\label{eq:foward_pass_mod}
    &x_{l+1,j} = \sum_{i=1}^{n_l} \tilde{x}_{l,j,i} = \sum_{i=1}^{n_l} \phi_{l,j,i}(x_{l,i})=\\ 
    &\sum_{i=1}^{n_l} \left( \omega^\varphi_{l,j,i} \cdot \varphi_{l,j,i}(x_{l,i}) + \omega^\mathbf{b}_{l,j,i} \cdot \mathbf{b}(x_{l,i}) \right) + \theta_{l,j}, \\
    &\forall \; j=1,\cdots,n_{l+1},
\end{aligned}
\end{equation}
where $\omega^\varphi_{l,j,i}$ and $\omega^\mathbf{b}_{l,j,i}$ are the weights associated respectively with the spline function and the base activation function connecting neuron $(l,i)$ to neuron $(l+1,j)$. Moreover, $\theta_{l,j}$ represents the bias added to neuron $(l+1,j)$. Besides, each spline $\varphi_{l,j,i}$ is given by the cubic Hermite spline of the data $\mathcal{X} = \{(x^{k}_{l,j,i}, y^{k}_{l,j,i}, m^{k}_{l,j,i}) \mid \forall \; 1 \leq k \leq K\}$ where $K$ is the number of knots.


Consequently, if $\omega^\varphi_{l,j,i}$ and $\omega^\mathbf{b}_{l,j,i}$ are positive and $\varphi_{l,j,i}$ and $\mathbf{b}$ are monotonic, for all $1 \leq i \leq n^l$, $1 \leq j \leq n^{l+1}$ and $0 \leq l \leq L-1$, then the resulting activation value function is also monotonic. Moreover, conditions established in Lemma \ref{teo:nec_condition_hermite} and Lemma \ref{teo:suf_condition_hermite} give us an intuitive way of imposing monotonicity for each of the splines $\varphi_{l,j,i}$.

Additionally, it is proposed that the cubic Hermite spline is extended linearly outside the interval of the definition of the spline $I = [x^1, x^K]$, with slope $m^1$ to the left of $x^1$ and slope $m^K$ to the right of $x^K$. This linear extrapolation ensures that each of the splines is \( C^1 \) continuous in  $\mathbb{R}$. Moreover, it also guarantees that the splines are monotonic, not just within the interval of definition, but in $\mathbb{R}$. Hence, the resulting KAN maintains monotonic consistency beyond the data domain and thereby certifies the monotonicity of the model across $\mathbb{R}^n$.

This idea is presented in Theorem \ref{teo:mono_conditions} that states the set of sufficient conditions needed to guarantee partial monotonicity of a KAN w.r.t the $r^{th}$ input. A detailed glossary of the notation used can be found in Appendix  \ref{ap:symbol_glossary}. Moreover, a complete proof of the above theorem can be found in Appendix \ref{sec:ap_mono_proof}.

\begin{theorem}[Certified Partial Monotonicity Conditions]\label{teo:mono_conditions}
Let $f: \mathbb{R}^n \rightarrow \mathbb{R}$ be a KAN composed of $L$ layers and $n$ inputs, where each spline function is defined over a common interval $I = [x^1, x^K]$ with $K$ knots. For each layer $l$, let $y^k_{l,j,i}$ denote the spline value at knot $k$ for the connection from neuron $i$ in layer $l$ to neuron $j$ in layer $l+1$. Let $d^k_{l,j,i} = (y^{k+1}_{l,j,i} - y^k_{l,j,i})/(x^{k+1}_{l,j,i} - x^k_{l,j,i})$ be the difference between the knot values, and let $m^k_{l,j,i}$ be the corresponding spline slope at knot $k$. Denote by $\omega^\varphi_{l,j,i}$ and $\omega^\mathbf{b}_{l,j,i}$ the weights applied to the spline and the base activation function output, respectively. Then, assuming the base activation function $\mathbf{b}$ is monotonically increasing, the KAN $f$ is certified increasingly (respectively decreasingly) monotonic with respect to the $r^{\text{th}}$ input variable if the following conditions are satisfied $\forall \; 1 \leq k \leq K$:

\medskip
\noindent\textbf{Input layer conditions for increasing (resp. decreasing) monotonicity with respect to the $r^{\text{th}}$ input:}
\begin{enumerate}
    \item $\omega^\varphi_{0,j,r} \geq 0$, $\omega^\mathbf{b}_{0,j,r} \geq 0$ (resp. $\omega^\varphi_{0,j,r}\geq 0, \omega^\mathbf{b}_{0,j,r}\leq 0 $) \hfill (non-negative weights on input $x_r$)
    \item $y^{k+1}_{0,j,r} \geq y^{k}_{0,j,r}$, i.e. $d^{k}_{0,j,r}\geq 0$ (resp. $y^{k+1}_{0,j,r}\leq y^{k}_{0,j,r}$, i.e. $d^{k}_{0,j,r}\leq 0$)  \hfill (monotonic spline values)
    \item If $d^k_{0,j,r} = 0$, then $m^k_{0,j,r} = m^{k+1}_{0,j,r} = 0$ \hfill (zero slope at flat intervals)
    \item If $d^k_{0,j,r} > 0$, then $m^k_{0,j,r} \geq 0$, $m^{k+1}_{0,j,r} \geq 0$ (resp. $\text{if} \;  d^{k}_{0,j,r} < 0 , \; m^{k}_{0,j,r}, \; m^{k+1}_{0,j,r}\leq 0$) \hfill (positive slopes)
    \item If $d^k_{0,j,r} > 0$, then $\left(\alpha^k_{0,j,r}\right)^2 + \left(\beta^k_{0,j,r}\right)^2 \leq 9$ \hfill (sufficient condition for monotonicity (Lemma 2))
\end{enumerate}

\medskip
\noindent\textbf{Hidden layers ($1 \leq l \leq L - 1$) increasing conditions:}
\begin{enumerate}
    \setcounter{enumi}{5}
    \item $\omega^\varphi_{l,j,i} \geq 0$, $\omega^\mathbf{b}_{l,j,i} \geq 0$ \hfill (non-negative weights)
    \item $y^{k+1}_{l,j,i} \geq y^{k}_{l,j,i}$ \hfill (monotonic spline values)
    \item If $d^k_{l,j,i} = 0$, then $m^k_{l,j,i} = m^{k+1}_{l,j,i} = 0$ \hfill (zero slope at flat intervals)
    \item If $d^k_{l,j,i} > 0$, then $m^k_{l,j,i}, m^{k+1}_{l,j,i} \geq 0$ \hfill (positive slopes at non flat intervals)
    \item If $d^k_{l,j,i} > 0$, then $\left(\alpha^k_{l,j,i}\right)^2 + \left(\beta^k_{l,j,i}\right)^2 \leq 9$ \hfill (sufficient condition for monotonicity (Lemma 2)).
\end{enumerate}
\end{theorem}

Lastly, it is worth mentioning that the proposed method, considering a linear combination with positive coefficients of monotonic functions, could be seen as analogous to \citep{Archer1993ApplicationProblems}, where a traditional MLP architecture with a ReLU activation function is constrained to have positive weights. However, the constraint proposed in \citep{Archer1993ApplicationProblems} significantly reduces the expressive power as it forces the output function to be convex \citep{Liu2020CertifiedNetworks}. In contrast, MonoKAN is capable of generating monotonic non-convex functions because it leverages monotonic cubic Hermite splines, which allow for flexible piecewise constructions and can model complex, non-convex shapes.  

\subsection{MonoKAN Algorithm}



Finally, we present the algorithm that guarantees a KAN satisfies the sufficient conditions defined in Theorem~\ref{teo:mono_conditions}, thereby certifying the network as partially monotonic. To enforce these constraints during training, we introduce a clamping mechanism that adjusts the learned parameters after each optimization step to ensure that the parameters remain within the valid range.

Although implemented as simple bound enforcement (i.e., clipping values to lie within predefined intervals), this procedure is mathematically equivalent to a projection onto a convex constraint set. As such, it can be interpreted as a special case of Projected Gradient Descent (PGD), where the projection ensures constraint satisfaction after each update \citep{Bubeck2015ConvexComplexity}. This projection step is performed independently of the gradient computation and does not alter the optimizer’s internal state, making it compatible with commonly used algorithms such as SGD and Adam.


Mathematically, consider $f: \mathbb{R}^n \to \mathbb{R}$, a KAN with $L$ layers, expected to be partially monotonic with respect to the $r^{th}$ input ($1 \leq r \leq n$). The sufficient conditions outlined in Theorem~\ref{teo:mono_conditions} are enforced through the clamping mechanism described in the \hyperref[ag:applyconstraints]{\textit{applyCons} Algorithm}, which is applied at every training epoch. As detailed in Section~\ref{sec:math_certification}, clamping must be applied to the spline activations directly influenced by $x_r$ and the downstream layers. A full pseudocode implementation is provided in the \hyperref[ag:monokan]{\textit{MonoKAN} Algorithm}.

\begin{algorithm}[!htp]
\caption{\textit{applyCons}}
\begin{algorithmic}[1]
\REQUIRE Parameters: 1-D arrays $\omega^\varphi_{l,:,i}$, $\omega^\mathbf{b}_{l,:,i}$ and 2-D arrays $y^k_{l,:,i}$, $m^k_{l,:,i}$, $x^k_{l,:,i}$ with $0 \leq k \leq K-1$.
\ENSURE Adjusted parameters fulfilling sufficient conditions from Theorem \ref{teo:mono_conditions} for $i^{th}$ input of layer $l$.
\STATE $\omega^\varphi_{l,:,i}, \;\omega^\mathbf{b}_{l,:,i} \leftarrow \max(0,\omega^\varphi_{l,:,i}), \;\max(0,\omega^\mathbf{b}_{l,:,i})$
\FOR{ $k$ in \text{range}($K-1$) }
    \STATE $y^{k+1}_{l,:,i} \leftarrow \max(y^{k+1}_{l,:,i} , y^{k}_{l,:,i})$ \COMMENT{Impose that the sequence of control points is increasing}
    \STATE $d^{k}_{l,:,i} \leftarrow \frac{y^{k+1}_{l,:,i} - y^{k}_{l,:,i}}{x^{k+1}_{l,:,i}- x^{k}_{l,:,i}}$
    \IF{$d^{k}_{l,:,i} = 0$}
        \STATE $m^{k}_{l,:,i}, m^{k+1}_{l,:,i} \leftarrow 0$
    \ELSE
        \STATE $m^{k}_{l,:,i}, m^{k+1}_{l,:,i} \leftarrow \max\left(0, m^{k}_{l,:,i}\right), \max\left(0, m^{k+1}_{l,:,i}\right)$ \COMMENT{Impose positivity of the derivatives}
    \STATE $\alpha^{k}_{l,:,i} \leftarrow \frac{m^{k}_{l,:,i}}{d^{k}_{l,:,i}}$
    \STATE $\beta^{k}_{l,:,i} \leftarrow \frac{m^{k+1}_{l,:,i}}{d^{k}_{l,:,i}}$
        \IF{$\left(\alpha^{k}_{l,:,i}\right)^2 + \left(\beta^{k}_{l,:,i}\right)^2 > 9 $}
            \STATE $\tau^{k}_{l,:,i} \leftarrow \frac{3}{\sqrt{(\alpha^{k}_{l,:,i})^2 + (\beta^{k}_{l,:,i})^2}}$
            \STATE $\alpha^{k}_{l,:,i} \leftarrow \tau^{k}_{l,:,i} \cdot \alpha^{k}_{l,:,i}$
            \STATE $\beta^{k}_{l,:,i} \leftarrow \tau^{k}_{l,:,i} \cdot \beta^{k}_{l,:,i}$
            \STATE $m^{k}_{l,:,i} \leftarrow \alpha^{k}_{l,:,i} \cdot d^{k}_{l,:,i}$
            \STATE $m^{k+1}_{l,:,i} \leftarrow \beta^{k}_{l,:,i} \cdot d^{k}_{l,:,i}$
            
        \ENDIF
    \ENDIF
\ENDFOR
\RETURN $\omega^\varphi_{l,:,i}$, $\omega^\mathbf{b}_{l,:,i}$, $y^k_{l,:,i}$ and $m^{k}_{l,:,i}$.
\end{algorithmic}\label{ag:applyconstraints}
\end{algorithm}

\begin{algorithm}[!htp]
\caption{\textit{MonoKAN Algorithm}}
\begin{algorithmic}[1]
\REQUIRE KAN model $f$ with $L$ layers and $K$ knots, maximum number of epochs \textit{max\_epochs}, index $r$ of the increasing (resp. decreasing) monotonic feature.
\ENSURE Adjusted parameters of the KAN to fulfill sufficient conditions from Theorem \ref{teo:mono_conditions}.
\FOR{$\text{epoch} = 1$ \TO $\textit{max\_epochs}$}
    \STATE Compute loss: $\mathcal{L} \leftarrow \textit{ComputeLoss}(f)$
    \STATE Perform optimizer step: $f \leftarrow \textit{OptimizerStep}(f, \mathcal{L})$
    \FOR{$l$ in range($L$)}
        \IF{$l = 0$}
            \STATE $\omega^\varphi_{0,:,r}, \omega^\mathbf{b}_{0,:,r},  y^k_{0,:,r},  m^k_{0,:,r} \leftarrow \textit{applyCons}(\omega^\varphi_{0,:,r}, \omega^\mathbf{b}_{0,:,r},  y^k_{0,:,r}, m^k_{0,:,r}, x_{0,:,r})$
            \STATE $\big(\text{resp.} \quad \omega^\varphi_{0,:,r}, -\omega^\mathbf{b}_{0,:,r},  -y^k_{0,:,r},  -m^k_{0,:,r} \leftarrow \textit{applyCons}(\omega^\varphi_{0,:,r},-\omega^\mathbf{b}_{0,:,r},  -y^k_{0,:,r}, -m^k_{0,:,r}, x_{0,:,r}) \big)$
        \ELSE
            \FOR{ $ i$ in range$\left(n^l\right)$}
             \STATE $\omega^\varphi_{l,:,i}, \omega^\mathbf{b}_{l,:,i},  y^k_{l,:,i},  m^k_{l,:,r} \leftarrow \textit{applyCons}(\omega^\varphi_{l,:,i}, \omega^\mathbf{b}_{l,:,i},  y^k_{l,:,i},  m^k_{l,:,r}, x_{l,:,i})$ 
             \ENDFOR
        \ENDIF
    \ENDFOR
\ENDFOR
\RETURN Partial monotonic KAN w.r.t the $r^{th}$ input
\end{algorithmic}\label{ag:monokan}
\end{algorithm}

\section{Experiments}\label{sec:experiments}

To assess the practical applicability of the proposed method, we divide our experimental analysis into two parts. First, we present a case study based on a two-dimensional setting from the \textit{ESL} dataset \citep{Ben-David1989LearningConcepts}, intended to provide intuitive insight into the behavior of MonoKAN and to illustrate the ethical importance of certified partial monotonicity. This case study aims to visually analyze  how unconstrained or partially constrained models can produce unfair or counterintuitive predictions, motivating the need for strict monotonicity guarantees in certain domains.

Next, we conduct a comprehensive set of experiments across multiple benchmark datasets to evaluate MonoKAN's performance relative to existing state-of-the-art models. Although no prior work has introduced a certified partial monotonic Kolmogorov–Arnold Network, we compare against leading monotonic MLP-based methods reported in the literature. 

To ensure a fair and consistent comparison, we adopted the experimental procedures established by \citet{Liu2020CertifiedNetworks} and \citet{Sivaraman2020Counterexample-GuidedNetworks}, which are widely accepted in the literature. Therefore, for each dataset, the experiments were conducted three times to report the mean and the standard deviation. Moreover, the dataset is divided using a three-way split, where 80$\%$ of the data is used for training and validation, and the remaining 20$\%$ for testing. Within the training portion, an additional 80$\%$/20$\%$ split is used to separate a validation set for early stopping. Consequently, although we benchmark our results against the state-of-the-art monotonic architectures \citep{Nolte2022ExpressiveNetworks,Runje2023ConstrainedNetworks}, we reran their experiments using the methodology of \citet{Liu2020CertifiedNetworks} and \citet{Sivaraman2020Counterexample-GuidedNetworks}. This approach ensures methodological consistency and allows for a fair comparison across the different studies. To support reproducibility and clarify the scope of our evaluation, we have added Appendix \ref{ap:data_description}, which contains detailed descriptions of all datasets used, including task type, input dimensionality, number of monotonic features, and the rationale behind the constraints applied. It is important to note that the monotonic feature selection criteria were not selected by the authors but are instead defined in the original benchmark studies \citep{Liu2020CertifiedNetworks, Sivaraman2020Counterexample-GuidedNetworks}.

All computations were carried out on a server equipped with two Intel(R) Xeon(R) Platinum 8480C CPUs. Each processor provides 56 physical cores and supports 2 threads per core, resulting in a total of 224 hardware threads across both sockets. The CPUs operate at a maximum clock speed of 3.8~GHz. The system is also provisioned with approximately 2.1~TB of RAM, ensuring sufficient memory for all experimental workloads. Moreover, for GPU-accelerated experiments, the server was provisioned with an NVIDIA H200 GPU (driver version 550.90.07, CUDA 12.4).
The proposed code was developed based on the Pytorch framework \citep{Paszke2019PyTorch:Library}.  The results and the proposed MonoKAN code can be accessed on GitHub at \href{https://github.com/alejandropolo/MonoKAN}{\url{https://github.com/alejandropolo/MonoKAN}}.



\subsection{Case Study: Visualizing Monotonicity and Fairness in 2D Predictions}

First of all, we consider the \textit{ESL} dataset \citep{Ben-David1989LearningConcepts}, which is one of the benchmarks in the literature of monotonic datasets \citep{Cano2019MonotonicSets}. The dataset consists of 488 records, each representing a job candidate evaluated for an industrial job. In particular, each of the records contains four input features representing the candidate's scores from standardized psychometric tests conducted during the assessment process. Besides, the output feature of the dataset is the alignment between the candidate's profile and the job requirements. 

Out of the four input variables, we decided to select the first two input features to consider a two-dimensional case study. Therefore, this 2D scenario serves as an ideal dataset to assess and visualize the proposed method and how the conditions presented in Theorem \ref{teo:mono_conditions}, lead to a certified partial monotonic KAN. Moreover, we decided to transform the output variable into a binary classification task by assigning a value of 1 to candidates with a suitability score greater than 5, and 0 otherwise. 

Given that the model's output is used to select candidates, ensuring fairness in the model's predictions is essential. In particular, a fair model should reflect an increase in a candidate’s predicted suitability for the job whenever there is an improvement in any of the psychometric test scores, assuming all other factors remain constant. Consequently, penalizing candidates for demonstrating stronger competencies undermines merit-based selection and can lead to unethical outcomes \citep{Hunter1976CriticalBias}. Specifically, a non-monotonic model may disadvantage more qualified applicants, producing hiring decisions that appear arbitrary or biased. Therefore, enforcing monotonicity with respect to each of these input features is essential to ensure that the model's behavior aligns with fundamental fairness principles.

Moreover, an important characteristic of this dataset is the uneven distribution of samples across the input space, which results in sparse coverage in certain regions. Such sparsity may lead unconstrained models to overfit or behave unreliably in less populated areas, producing erratic or unintuitive predictions. This reinforces the relevance of incorporating certified monotonic constraints to ensure model stability and trustworthy behavior.

To evaluate the impact of certified partial monotonicity on model behavior and performance, we trained four variants of KANs on the two-dimensional ESL dataset: an unconstrained KAN using the original B-splines, an unconstrained KAN using the proposed cubic Hermite splines, a partially constrained MonoKAN (monotonic with respect to the first psychometric test), and a fully constrained MonoKAN (monotonic with respect to both input features). Table~\ref{tab:esl_accuracy} reports the training and test accuracy for all four models. Notably, all models achieve nearly identical performance on the training and test sets. This suggests that enforcing monotonicity, whether partially or fully, does not appear to compromise predictive accuracy in this setting, as exploratory data analysis supports a monotonic relationship between inputs and outputs.

\begin{table}[htbp]
\centering
\caption{Accuracy on the ESL dataset under different monotonicity constraints}
\label{tab:esl_accuracy}
\setlength{\tabcolsep}{3pt}
\renewcommand{\arraystretch}{1.0}
\begin{tabular}{lcc}
\toprule
\textbf{Model} & \textbf{Train} & \textbf{Test} \\
\midrule
Unconstrained Hermite KAN & 0.80  & 0.85 \\
Unconstrained B-splines KAN & 0.80  & 0.85 \\
MonoKAN (1 feature monotonic) & 0.79  & 0.85 \\
MonoKAN (2 features monotonic) & 0.79  & 0.85 \\
\bottomrule
\end{tabular}
\end{table}

\begin{figure*}[h!]\
\centering     
\subfigure[]{
\includegraphics[width=37mm]{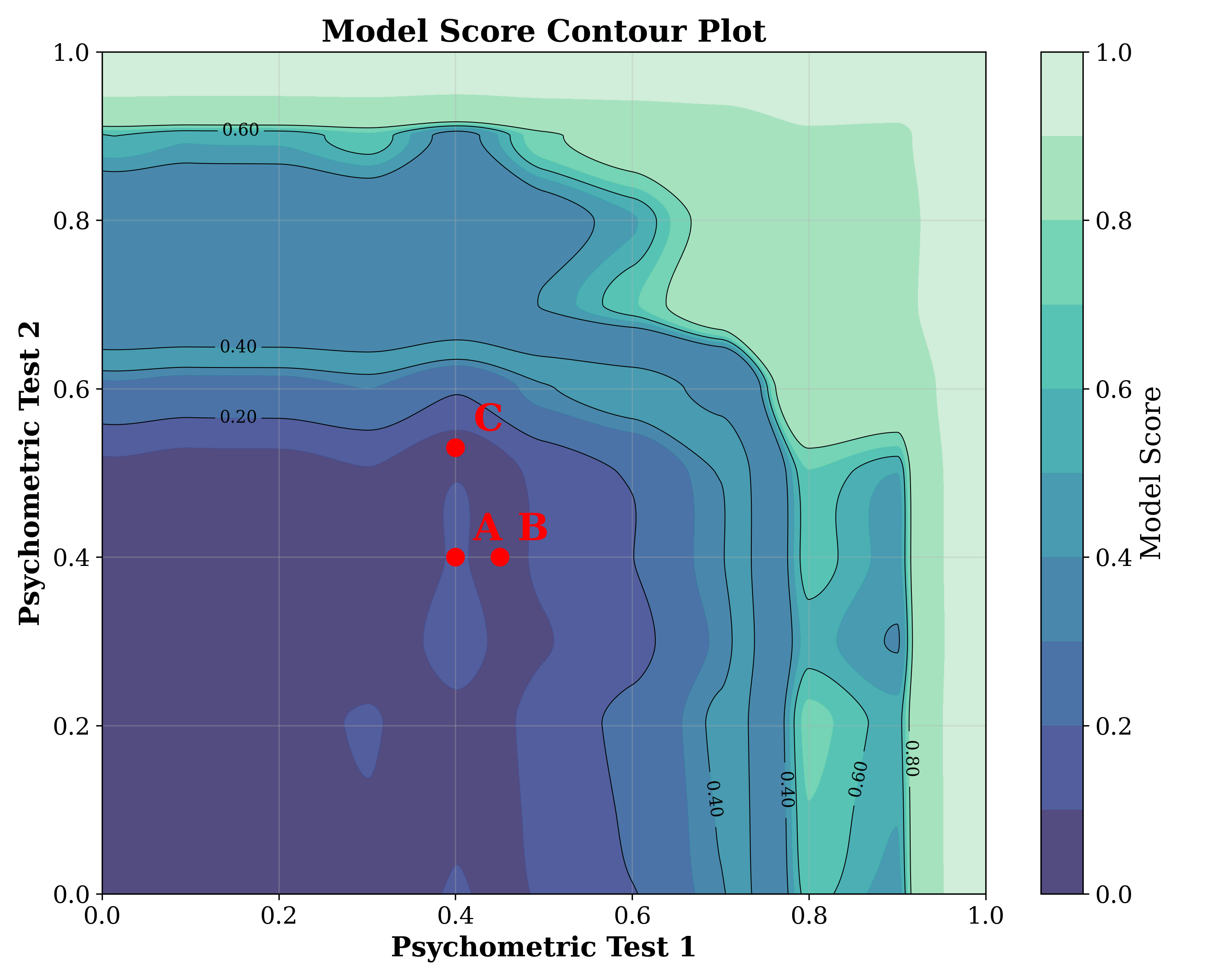}
}
\subfigure[]{
\includegraphics[width=37mm]{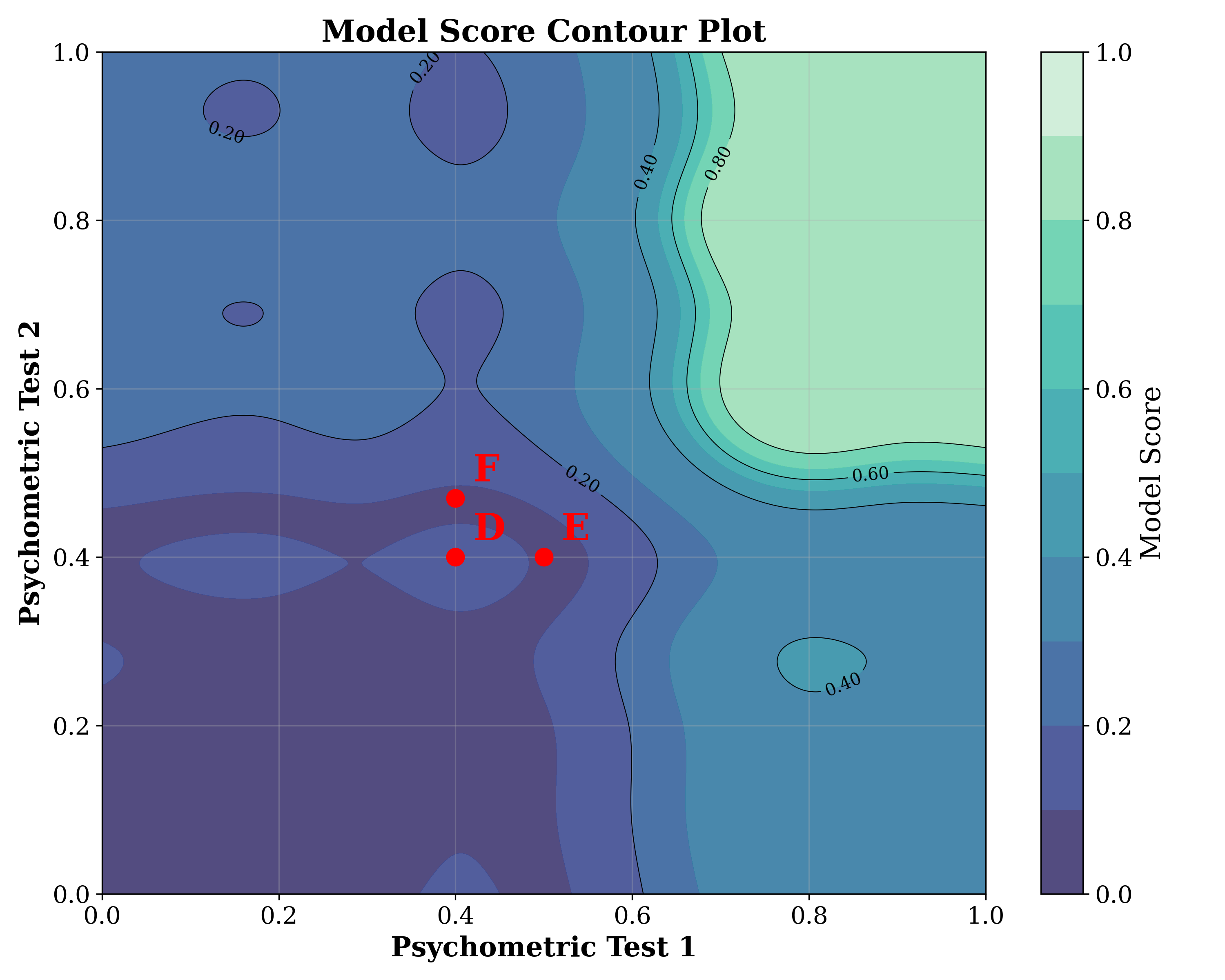}
}
\subfigure[]{
\includegraphics[width=37mm]{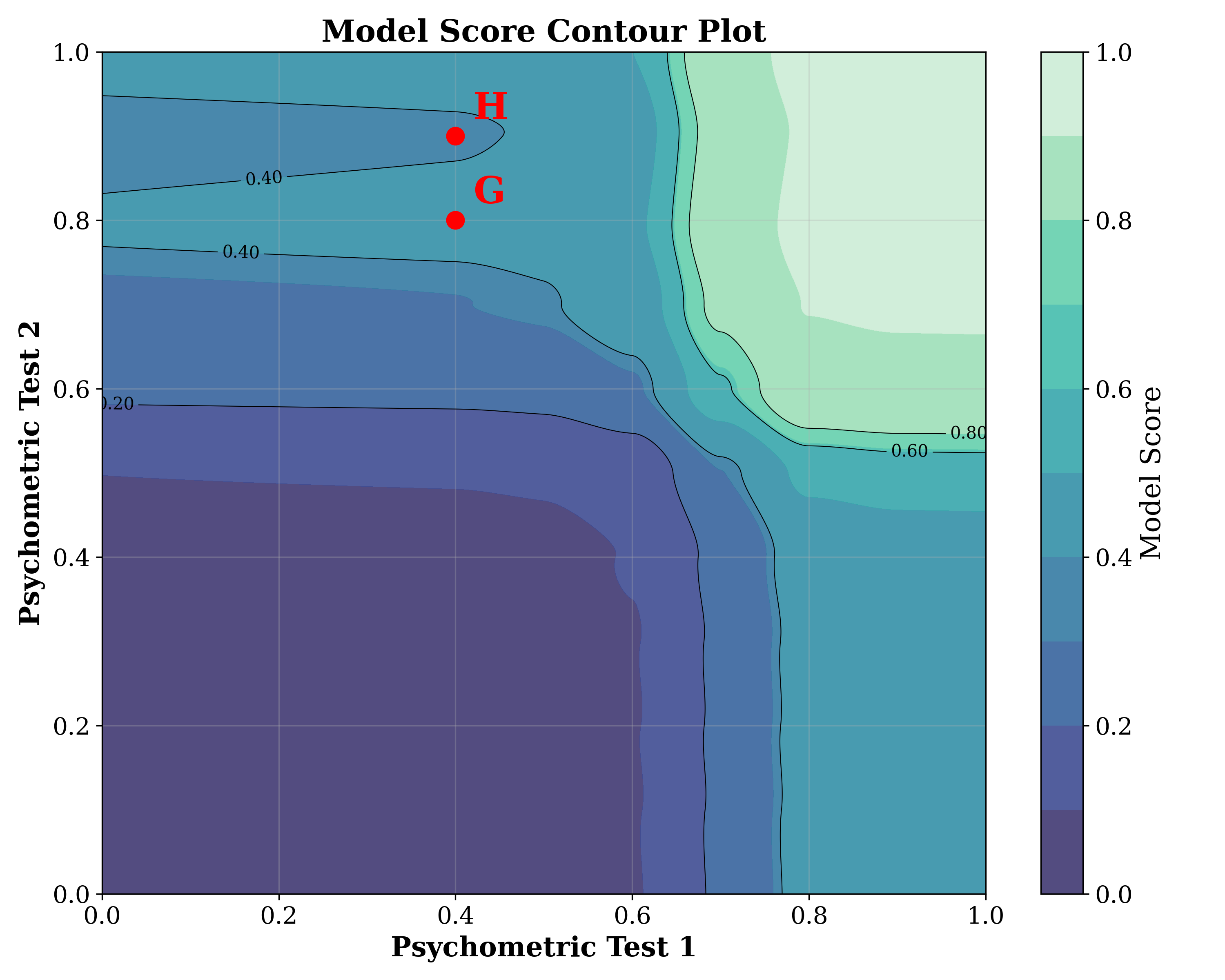}
}
\subfigure[]{
\includegraphics[width=37mm]{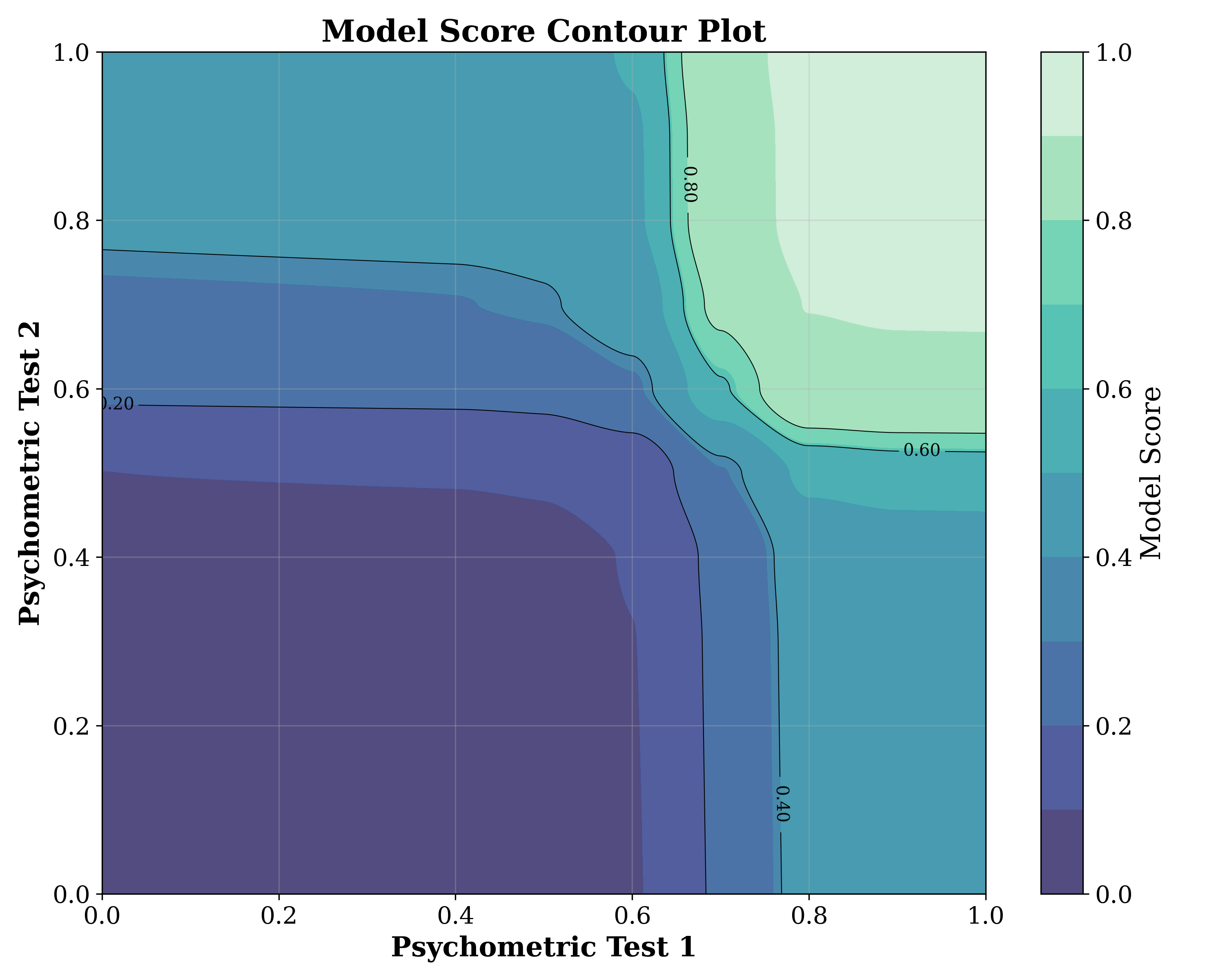}
}
\caption{Contour plots of model outputs over the 2D input space (psychometric test 1 vs. psychometric test 2) for (a) unconstrained cubic Hermite KAN, (b) unconstrained B-spline KAN, (c) partially constrained MonoKAN (monotonic with respect to the first psychometric test), and (d) fully constrained MonoKAN. In (a) and (b), the model exhibits several violations of monotonicity, as highlighted by candidate comparisons (e.g., A vs. B, and A vs. C), where improved input scores lead to lower predicted suitability. In (c), monotonicity is enforced only along test 1, improving consistency along that axis but still allowing irregularities along test 2 (G vs H). In (d), full monotonicity constraints result in a smooth, fair decision surface aligned with ethical expectations.}
\label{fig:contour_2d}
\end{figure*}

Although the accuracy scores across all four models are nearly identical, the key distinction lies in the fairness and monotonicity of their predictions, as illustrated in Figure~\ref{fig:contour_2d}. Both unconstrained models (Figure~\ref{fig:contour_2d} (a),(b)) displays several violations of monotonicity: increasing the score in either psychometric test 1 or test 2 can lead to a lower predicted suitability score. For example, considering Figure~\ref{fig:contour_2d} (a), comparing the selected candidates A and B, we observe that although candidate B has a higher score on psychometric test 1 and an equal score on test 2, the model assigns a higher suitability score to candidate A, demonstrating a counterintuitive and unfair outcome. A similar issue is observed between candidates A and C: both have the same score on psychometric test 1, but candidate C has a higher score on test 2, yet receives a lower prediction. The same behaviour can be easily identified in the case of the unconstrained KAN using the B-splines formulation. Consequently, these instances reveal how unconstrained models may fail to uphold basic fairness expectations in high-stakes contexts such as hiring.

On the other hand, the partially constrained model (Figure~\ref{fig:contour_2d} (c)), which enforces monotonicity only with respect to psychometric test 1, successfully eliminates violations along that axis. However, it still produces irregularities with respect to test 2. This is evident in the region around candidates G and H, where higher values in test 2 still lead to lower predicted outcomes, indicating that partial monotonicity is insufficient when fairness must be guaranteed in all relevant dimensions. In contrast, the fully constrained MonoKAN model (Figure~\ref{fig:contour_2d} (d) enforces certified monotonicity with respect to both inputs, resulting in a smooth and consistent decision surface. The model output increases as either psychometric test score improves, ensuring alignment with ethical expectations and eliminating unfair behaviors observed in the other configurations.

\begin{figure*}[h!]\
\centering     
\subfigure[]{
\includegraphics[width=37mm]{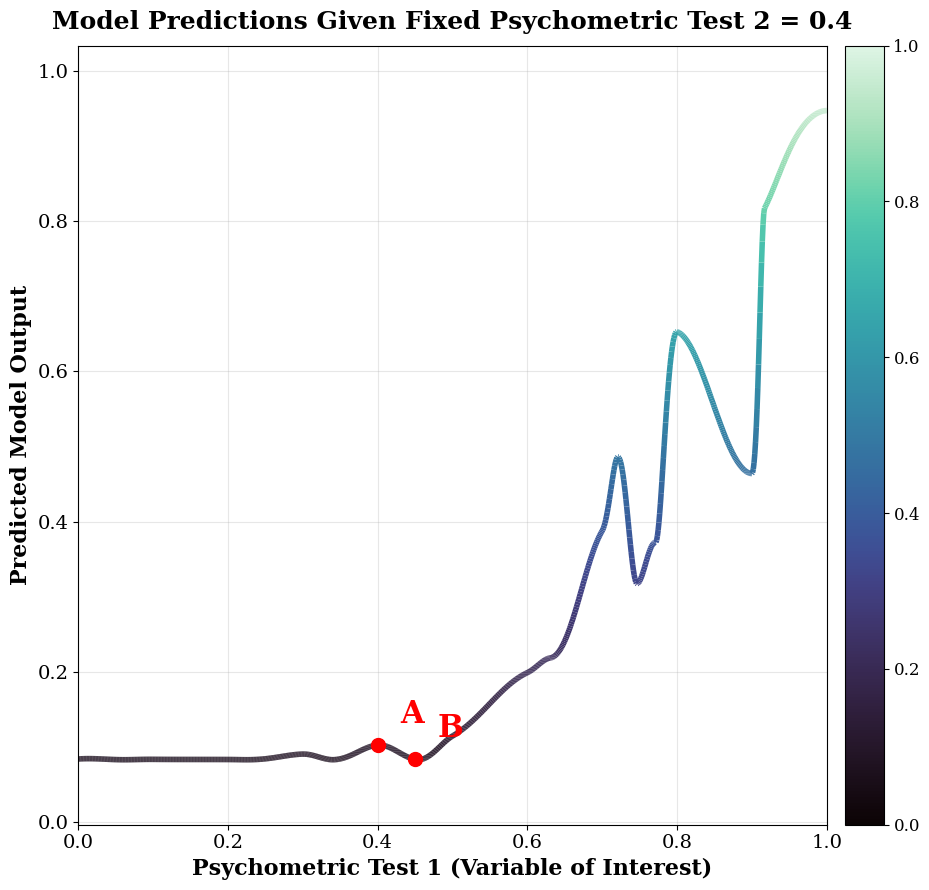}
}
\subfigure[]{
\includegraphics[width=37mm]{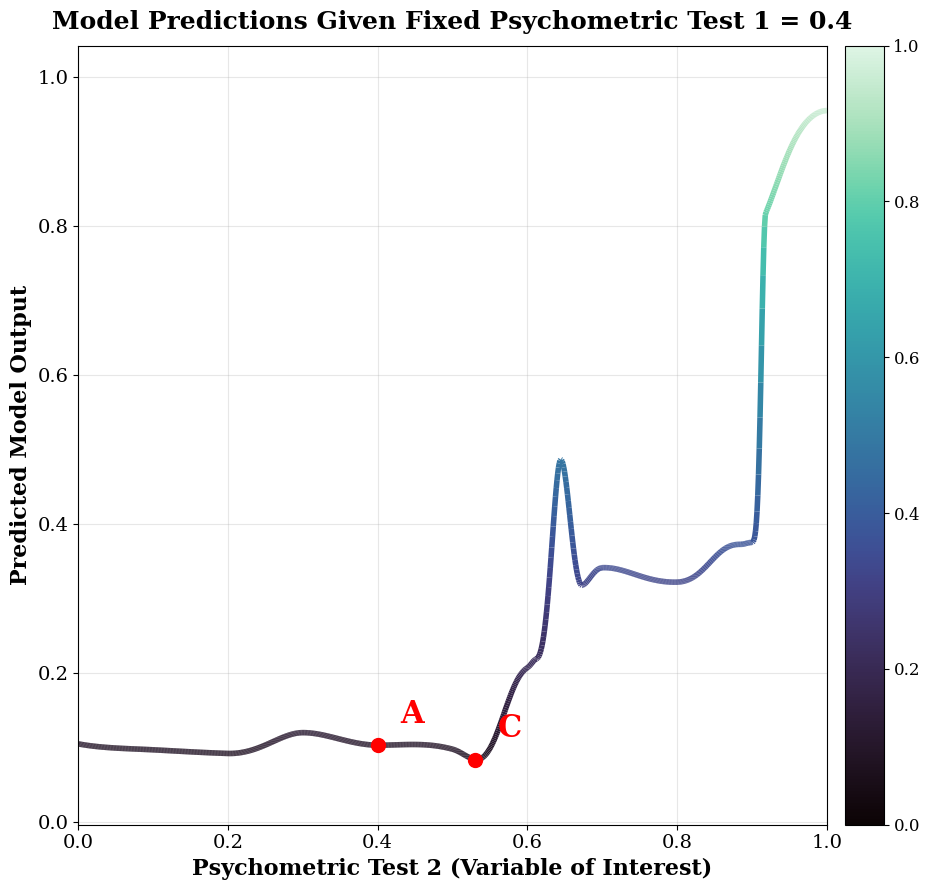}
}
\subfigure[]{
\includegraphics[width=37mm]{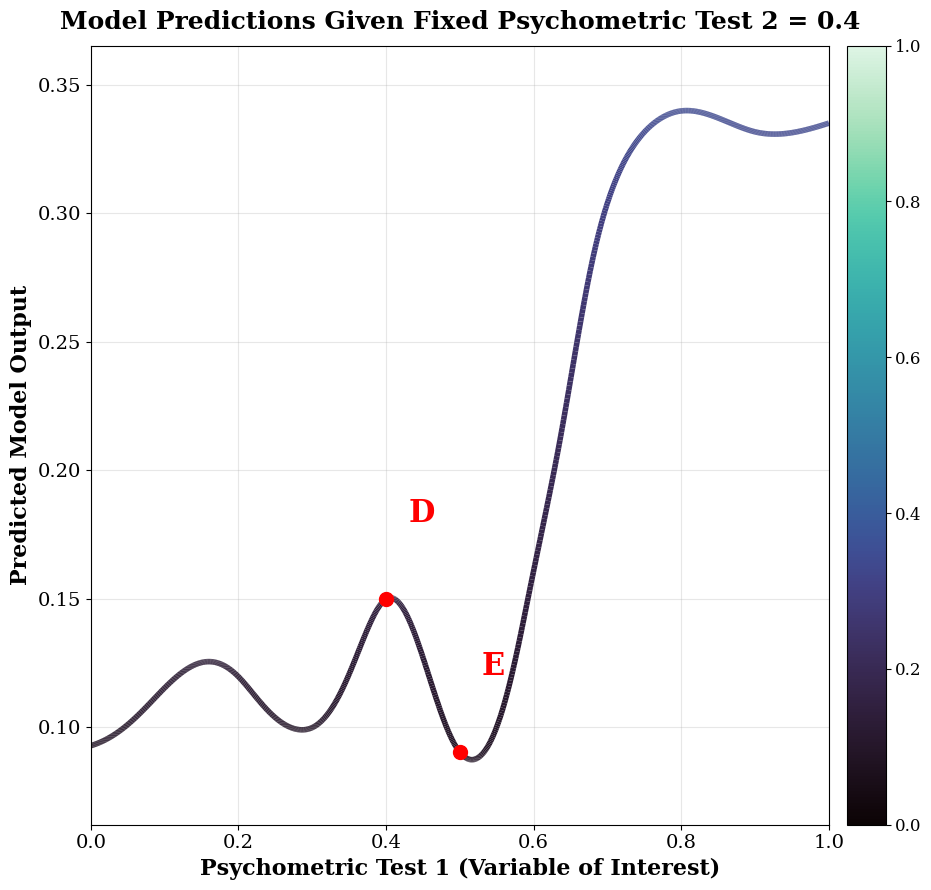}
}
\subfigure[]{
\includegraphics[width=37mm]{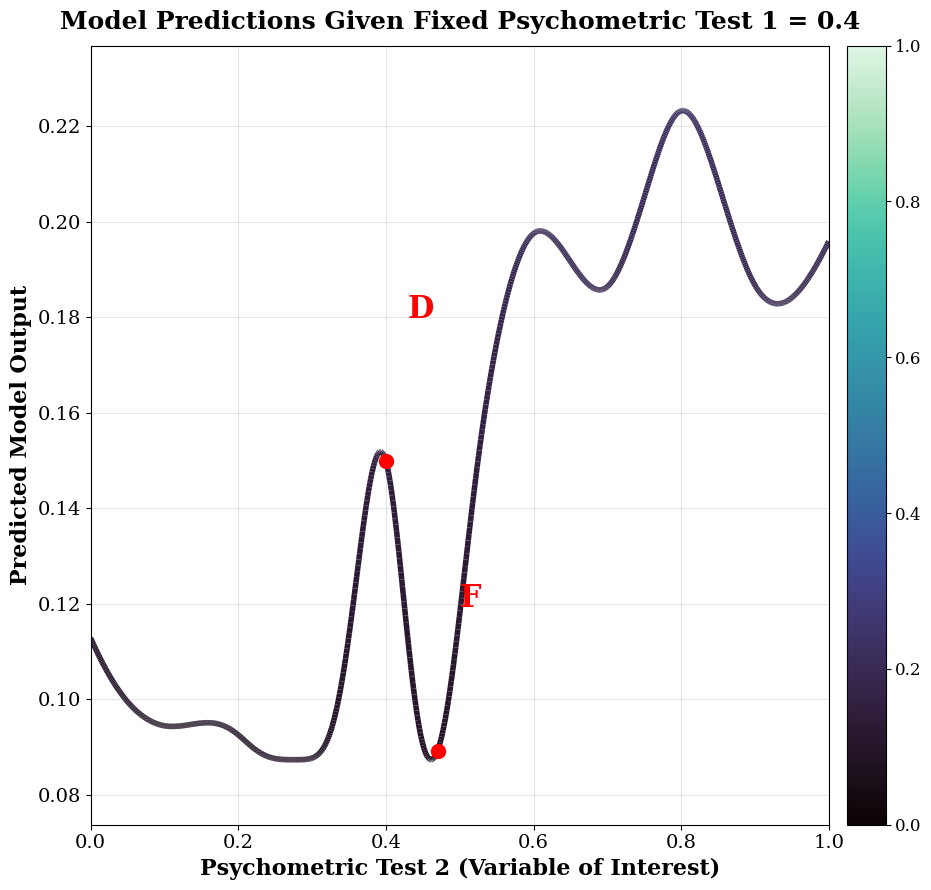}
}
\subfigure[]{
\includegraphics[width=37mm]{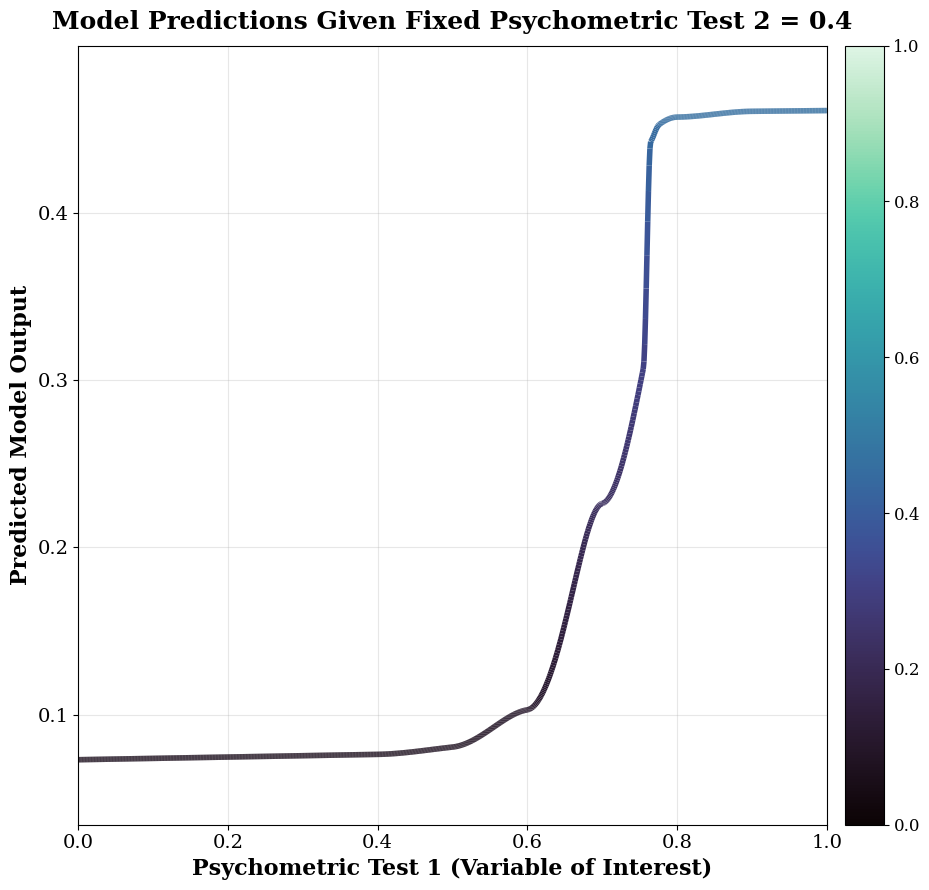}
}
\subfigure[]{
\includegraphics[width=37mm]{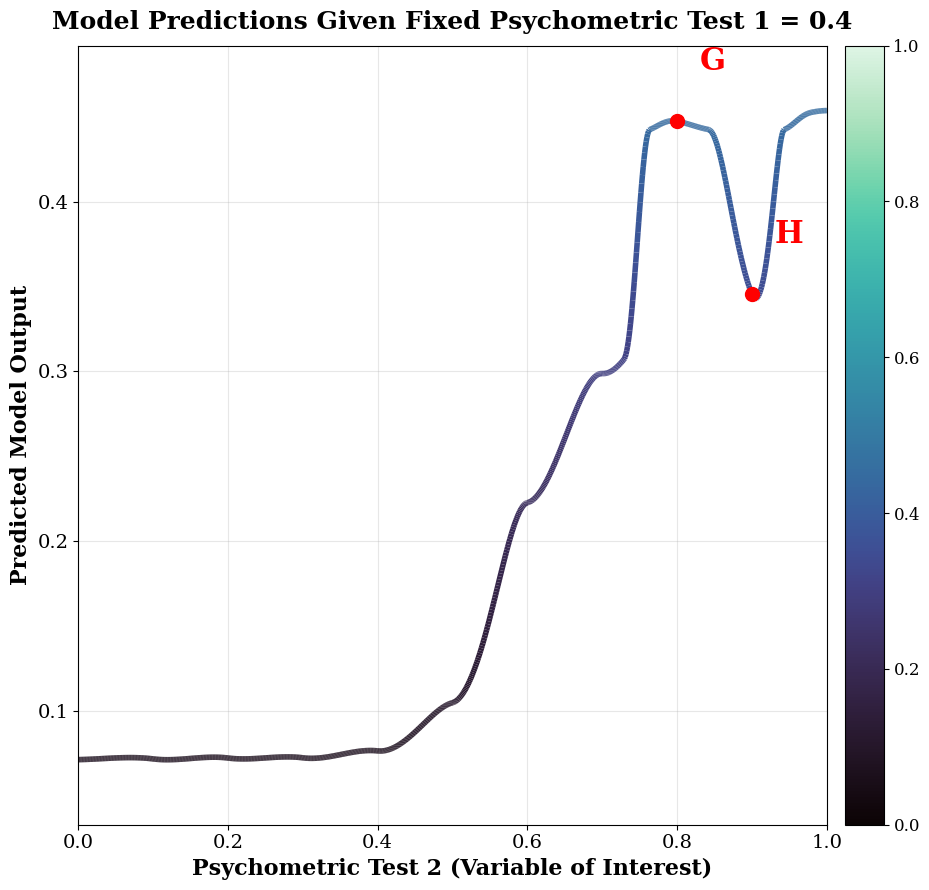}
}
\subfigure[]{
\includegraphics[width=37mm]{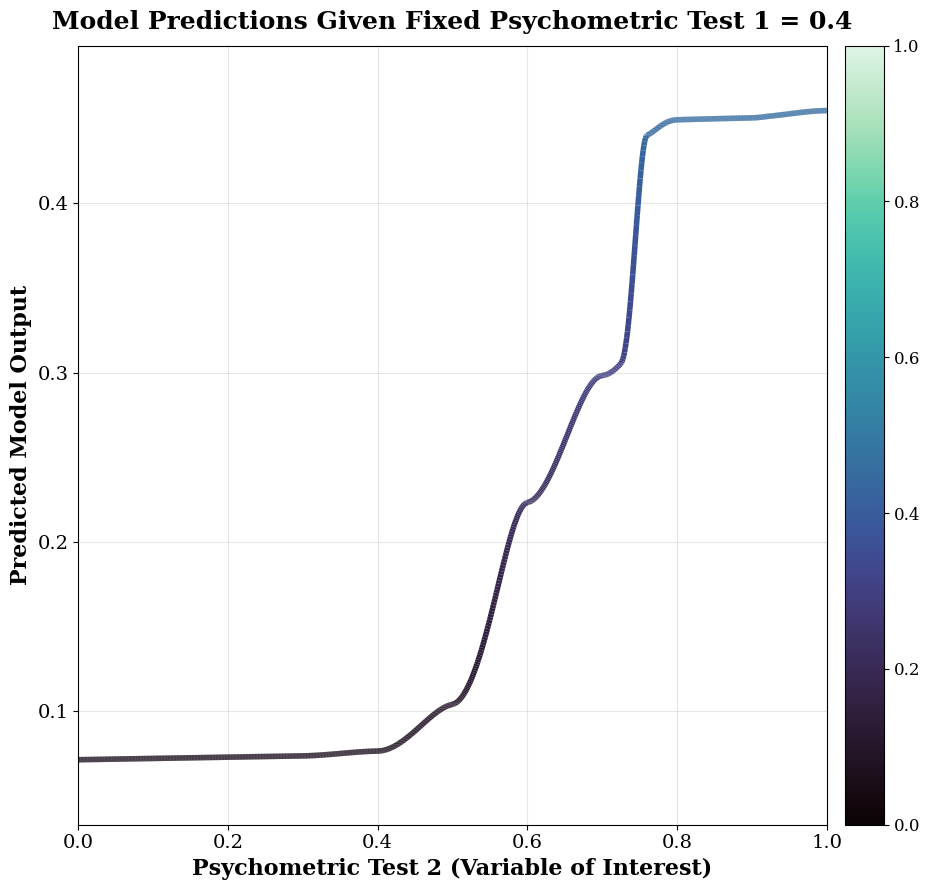}
}
\subfigure[]{
\includegraphics[width=37mm]{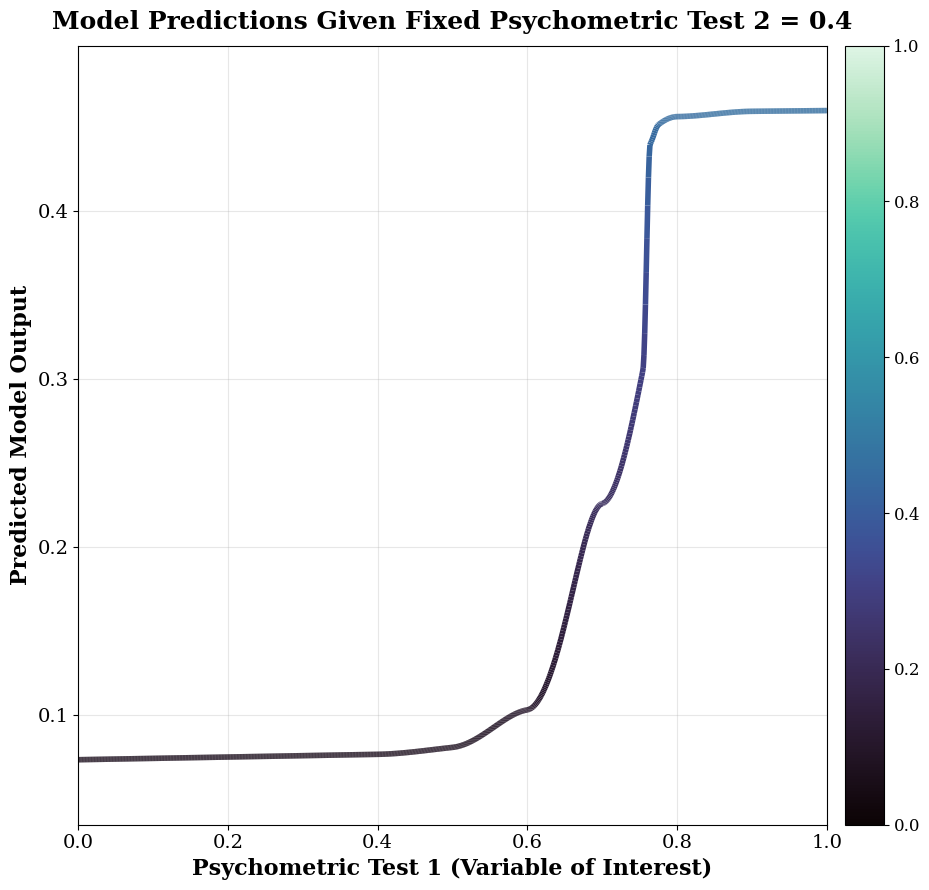}
}
\caption{One-dimensional slices of the model outputs across psychometric test 1 and test 2, with the complementary variable fixed at 0.4, to visualize local monotonicity violations. Each row corresponds to a different model architecture: (a, b) unconstrained cubic Hermite KAN, (c, d) unconstrained B-spline KAN, (e, f) partially constrained MonoKAN (monotonic with respect to the first psychometric test), and (g, h) fully constrained MonoKAN. Red dots highlight specific counterexamples (e.g., A vs. B, D vs. E, G vs. H) where increasing an input leads to a decrease in the predicted output, violating monotonicity. These examples are consistent with the patterns observed in the contour plots of Figure~\ref{fig:contour_2d}, providing further evidence of undesirable behaviors in unconstrained models. Notably, all violations are eliminated in the fully constrained MonoKAN (g, h), confirming its certified monotonic behavior across both input dimensions.}

\label{fig:slice_cuts}
\end{figure*}
To better illustrate these monotonicity violations at a finer granularity, Figure~\ref{fig:slice_cuts} presents one-dimensional slices of the prediction's surface of each of the aforementioned trained models, fixing one input while varying the other along the minimum and the maximum value of the dataset. These plots directly correspond to the candidate comparisons highlighted in Figure~\ref{fig:contour_2d}, allowing us to isolate how model predictions behave along each axis. For instance, in Figures~\ref{fig:slice_cuts}(a)--(d), clear local drops in predicted suitability are observed when increasing one psychometric test score while holding the other fixed, confirming the violations highlighted in the contour plots. These non-monotonic trends are particularly evident around points A–B or A-C and D–E, D-F, emphasizing the ethical risks of using unconstrained models. In the partially constrained MonoKAN model, shown in Figures~\ref{fig:slice_cuts}(e)--(f), we observe that no monotonicity violations occur when psychometric test 1 is varied (Figure~\ref{fig:slice_cuts}(e)), as constraints are imposed along this dimension. However, when psychometric test 2 is varied (Figure~\ref{fig:slice_cuts}(f)), the model still exhibits non-monotonic behavior, e.g., a drop between points G and H, highlighting that enforcing monotonicity in only one input dimension may be insufficient in fairness-critical applications. Finally, these irregularities are fully eliminated in Figures~\ref{fig:slice_cuts}(g)--(h), where the fully constrained MonoKAN produces strictly non-decreasing outputs along both inputs. This further supports the effectiveness of certified monotonicity in enforcing reliable and interpretable model behavior.

\begin{figure}[h]
\centering
\subfigure[]{
\includegraphics[width=37mm]{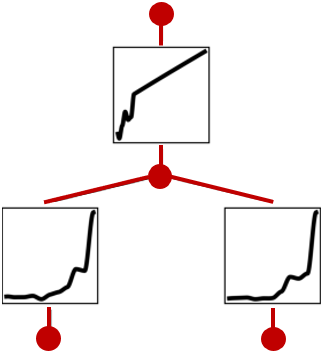}
}
\subfigure[]{
\includegraphics[width=37mm]{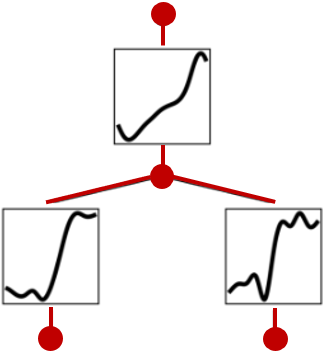}
}
\subfigure[]{
\includegraphics[width=37mm]{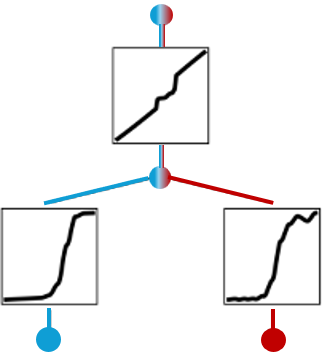}
}
\subfigure[]{
\includegraphics[width=37mm]{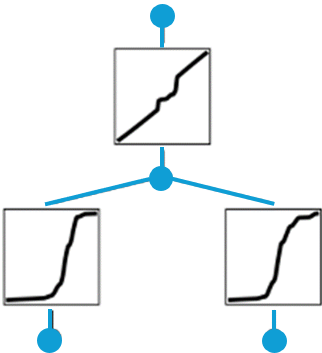}
}
\caption{Spline activation functions learned by the three models: (a) unconstrained cubic Hemite KAN, (b) unconstrained B-spline KAN, (c) partially constrained MonoKAN (monotonic with respect to psychometric test 1), and (c) fully constrained MonoKAN (monotonic with respect to both input features). The black curves represent the learned univariate spline activations. The unconstrained model exhibits non-monotonic shapes while the partially constrained model exhibits non-monotonic spline activations in the second input feature. Finally, in the certified monotonic MonoKAN all activations follow monotonic patterns.}
\label{fig:spline_act}
\end{figure}

Finally, to understand how these behaviors emerge, Figure~\ref{fig:spline_act} shows the learned spline activations for each neuron. In both unconstrained models, the spline activations exhibit local non-monotonicities that contribute to the counterintuitive predictions. These irregularities are partially corrected in the partially constrained model, and fully eliminated in the certified MonoKAN, where all activations conform to the required monotonicity constraints.

\subsection{Benchmarking MonoKAN Against State-of-the-Art Algorithms}

After the illustrative example presented in the previous section, we now evaluate the performance of the proposed algorithm on several real-world datasets. 
In the initial set of experiments, we used the following datasets proposed by \citep{Liu2020CertifiedNetworks}: COMPAS \citep{Angwin2016MachineBlacks.}, a classification dataset containing 6{,}172 records and 13 features, including 4 features labeled as monotonic; Blog Feedback Regression \citep{Buza2014FeedbackBlogs}, a high-dimensional regression dataset with 54{,}270 samples and 276 features, 8 of which are designated monotonic; and Loan Defaulter, a large-scale classification dataset with nearly half a million records and 28 features, including 5 monotonic ones \footnote{https://www.kaggle.com/datasets/wordsforthewise/lending-club/code}. Therefore, the aforementioned datasets enable a realistic evaluation of MonoKAN in real-world, large-scale scenarios that involve high-dimensional feature spaces.


Moreover, following the approach described in \citep{Nolte2022ExpressiveNetworks}, we evaluated both MonoKAN and the Expressive Monotonic Network using two configurations: one with the full set of input features and another with a reduced subset selected via Ridge regression. In the reduced setup, we trained a Ridge model and selected the top 20 features for the Blog Feedback dataset and the top 15 for the Loan Defaulter dataset, based on the absolute value of the learned coefficients. For each model, we reported the results corresponding to the best-performing configuration.

To improve clarity and reproducibility, we have added Appendix \ref{ap:data_description}, which provides detailed descriptions of each dataset, including task type, number of features, number of monotonic constraints, and the rationale behind the monotonic assignments. Besides, it is important to note that the monotonic features, described in Appendix \ref{ap:data_description}, were defined according to the original benchmark specifications and were not determined by the authors.

On the other hand, for the second set of experiments, we employed datasets specified by \citet{Sivaraman2020Counterexample-GuidedNetworks}: the Auto MPG dataset \footnote{https://archive.ics.uci.edu/dataset/9/auto+mpg}, which is a regression dataset with 3 monotonic features and is one of the benchmarks in the literature \citep{Cano2019MonotonicSets}, and the Heart Disease dataset \footnote{https://archive.ics.uci.edu/dataset/45/heart+disease}, which is a classification dataset featuring two monotonic variables. The results obtained are going to be compared with COMET \citep{Sivaraman2020Counterexample-GuidedNetworks}, Min-Max Net \citep{Daniels2010MonotoneNetworks}, Deep Lattice Network \citep{You2017DeepFunctions}, Constrained Monotonic Networks \citep{Runje2023ConstrainedNetworks} and \citep{Nolte2022ExpressiveNetworks}.

\subsubsection{Results}

The obtained results after performing the experiments are summarized in Tables \ref{tab:compas} and \ref{tab:comet}. As observed, the results obtained by the proposed MonoKAN outperform the state-of-the-art in four out of five datasets, while in the remaining experiment, our method equals the best option. 

When considering the number of parameters for each model, KAN remains competitive compared to most of the approaches proposed in the literature. However, in cases where the number of input variables is substantial, KAN exhibits a higher parameter count than some approaches. This increase in parameters for datasets with numerous inputs is due to KAN's architecture, which generates at least one spline in the first layer for each input. Consequently, the model complexity and the number of parameters grow proportionally with the number of input variables, impacting the overall efficiency and computational requirements of KANs for datasets with a large number of variables. 

\begin{figure}[!t]
\centering
\includegraphics[width=\linewidth]{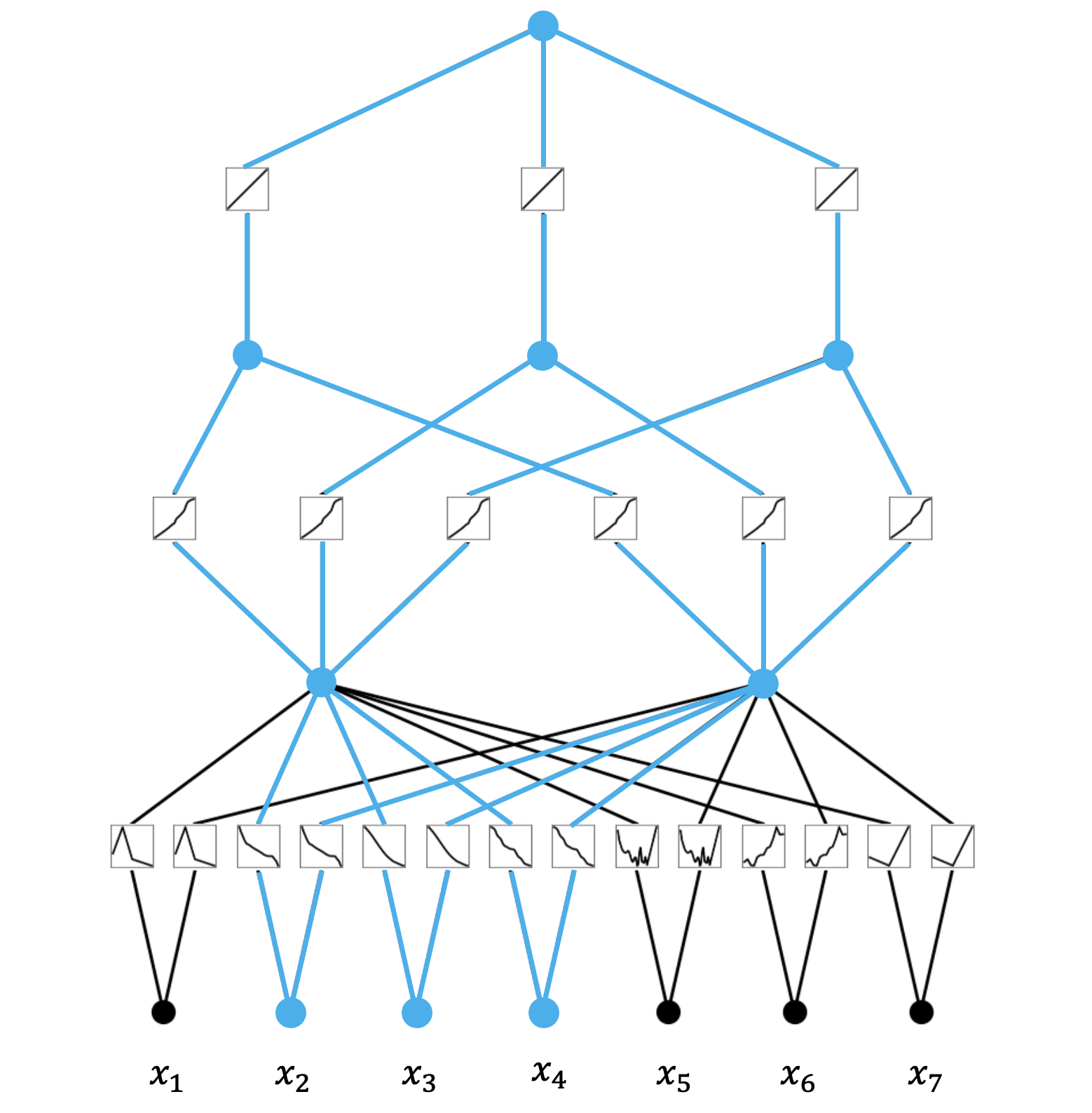}
\caption{Spline activations of a trained decreasing partial monotonic MonoKAN w.r.t the input variables $x_2,x_3$ and $x_4$ using the Auto MPG dataset.}
\label{fig:monokan_autompg}
\end{figure}

On the other hand, it is important to note that MonoKAN inherits all the additional advantages of using KAN architectures compared with traditional MLPs described in the introduction, especially its enhanced interpretability arising from being easier to visualize. For instance, in Figure \ref{fig:monokan_autompg}, we can observe a trained MonoKAN model using the Auto MPG dataset. This illustration highlights the specific relationships between each input variable and the output, demonstrating the certified decreasing partial monotonicity concerning input variables $x_2$, $x_3$, and $x_4$. The visualization effectively showcases how the MonoKAN model maintains monotonic behavior w.r.t these selected features. This is crucial for understanding and verifying the model's adherence to monotonic constraints, which can be essential for applications requiring reliable and interpretable predictions. Additionally, MonoKAN provides insight into the model's behavior and performance, allowing for a deeper analysis of the variable interactions and their impact on the output.

\begin{table*}[htbp]
\centering
\setlength{\tabcolsep}{2pt}
\renewcommand\arraystretch{1.3}
\small{
\begin{tabular}{l||rl|rl|rl}
\multirow{2}{*}{\bfseries Method} &
\multicolumn{2}{c|}{COMPAS}  &
\multicolumn{2}{c|}{Blog Feedback}  &
\multicolumn{2}{c}{Loan Defaulter} \\\cline{2-7}
   & Parameters & Test Acc  &
  Parameters & RMSE 
& Parameters & Test Acc  \\\hline
 Isotonic& N.A. & 67.6\% & N.A. & 0.203  & N.A. & 62.1\% \\
 XGBoost~\citep{Chen2016XGBoost:System}  & N.A. & 68.5\% $\pm$ 0.1\%  & N.A. & 0.176 $\pm$ 0.005  & N.A. & 63.7\% $\pm$ 0.1\%\\
 Crystal~\citep{MilaniFard2016FastLattices} & 25840 & 66.3\% $\pm$ 0.1\%  & 15840 & 0.164 $\pm$ 0.002 & 16940 & 65.0\% $\pm$ 0.1\% \\
 DLN~\citep{You2017DeepFunctions}     & 31403 & 67.9\% $\pm$ 0.3\% & 27903 & 0.161 $\pm$ 0.001& 29949  & 65.1\% $\pm$ 0.2\% \\
 Min-Max Net~\citep{Daniels2010MonotoneNetworks}     & 42000  & 67.8\% $\pm$ 0.1\% & 27700 & 0.163 $\pm$ 0.001  & 29000 & 64.9\% $\pm$ 0.1\%\\
  Non-Neg-DNN & 23112 & 67.3\% $\pm$ 0.9\% & 8492  & 0.168 $\pm$ 0.001 & 8502 & 65.1\% $\pm$ 0.1\%  \\
 Certified~\citep{Liu2020CertifiedNetworks} & 23112 & 68.8\% $\pm$ 0.2\%  & 8492 & 0.158 $\pm$ 0.001   & 8502 &  65.2\% $\pm$ 0.1\%  \\
 Constrained~\citep{Runje2023ConstrainedNetworks}   & 2317 & 65.3\%  $\pm$ 0.0\% & 1101 & 0.156 $\pm$ 0.001 & 577 & 65.0\% $\pm$ 0.0\%\\
 Expressive~\citep{Nolte2022ExpressiveNetworks}   & 37 & 68.7\% $\pm$ 0.0\% & 177 & 0.158$^1$ $\pm$ 0.003 & 753 & \textbf{65.3}\% $\pm$ 0.0 \% \\
 MonoKAN    & 4101 & \textbf{69.0\%} $\pm$ 0.0\% & 8546 & \textbf{0.155}$^1$ $\pm$ 0.0 & 1926 & \textbf{65.3}\%$^1$ $\pm$ 0.1 \%\\
\end{tabular}}
\vspace{5pt}
\caption{Comparison of the proposed MonoKAN with the state-of-the-art certified partial monotonic MLPs.}
\vspace{-\baselineskip} 
\begin{flushleft}
\footnotesize{$^1$ Following \citep{Nolte2022ExpressiveNetworks}, we apply Ridge regression-based feature selection, using the top 20 features for Blog Feedback and top 15 for Loan Defaulter}
\end{flushleft}
\label{tab:compas}
\end{table*}

\begin{table*}[htbp]
\centering
\setlength{\tabcolsep}{2pt}
\renewcommand\arraystretch{1.3}
\small{
\begin{tabular}{l||c|c}
\multirow{1}{*}{\bfseries Method} &
\multicolumn{1}{c|}{Auto MPG}  &
\multicolumn{1}{c}{Heart Disease} \\\cline{2-3}
& MSE  
 & Test Acc  \\\hline
  Min-Max Net~\citep{Daniels2010MonotoneNetworks}   & 10.14 $\pm$ 1.54  & 0.75 $\pm$ 0.04\\
 DLN~\citep{You2017DeepFunctions} & 13.34  $\pm$ 2.42& 0.86 $\pm$ 0.02 \\
 COMET~\citep{Sivaraman2020Counterexample-GuidedNetworks} & 8.81 $\pm$ 1.81  &  0.86 $\pm$ 0.03  \\
 Constrained ~\citep{Runje2023ConstrainedNetworks}    & 9.05 $\pm$ 0.25    & 0.86 $\pm$ 0.01 \\ 
 Expressive ~\citep{Nolte2022ExpressiveNetworks}    & 6.23 $\pm$ 0.50    & 0.81 $\pm$ 0.03 \\ 
 MonoKAN  & \textbf{6.18 $\pm$ 0.02}    & \textbf{0.89 $\pm$ 0.00} \\ 
\end{tabular}}
\vspace{5pt}
\caption{Comparison of the proposed MonoKAN with the state-of-the-art certified partial monotonic MLPs.}
\label{tab:comet}
\end{table*}

\section{Computational Implementation Details}\label{sec:computational_implementation_details}
This section provides a detailed analysis of the computational characteristics of the proposed MonoKAN architecture, including parameter complexity, memory usage, runtime behavior, and scalability considerations.

To begin with, regarding parameter complexity, a KAN with depth $L$ and width $N$ requires $O(N^2L(G + k))$ parameters, where $G$ represents the number of spline grid points and $k$ is the spline degree \citep{Liu2024KANScience, Kundu2024KANQAS:Search}. In our proposed MonoKAN framework, we utilize cubic Hermite splines resulting in a parameter complexity of $O(N^2L(2G))$, reamining comparable with the original KAN formulation. Hence, switching from B-splines to Hermite splines does not introduce significant differences in parameter complexity.

In comparison, a conventional MLP with the same architecture would require $O(N^2L)$ parameters \citep{Kundu2024KANQAS:Search}. Although this suggests that KANs and MonoKAN introduce an increase in parameter count, empirical studies show that KANs often require half the number of parameters than MLPs to achieve comparable performance \citep{Liu2024KAN:Networks, Tran2024ExploringImplementation}. This efficiency stems from the greater expressiveness of spline-based activations, which allows KAN-based models to achieve similar or better performance with reduced depth or width.

However, a key challenge in KANs is their dense connectivity: for a KAN model with $n$ input features and a first hidden layer with $n_1$ neurons, the network must learn $n \cdot n_1$ univariate spline functions in the first layer, since each input is connected to every neuron through a separate spline.  Therefore, this formulation can be inefficient in high-dimensional settings. To address this, sparse activation schemes have been proposed, where each neuron connects to only a subset of input features, thereby reducing the number of spline modules and memory overhead \citep{Liu2024KANScience}. Furthermore, MultKAN \citep{Liu2024KANScience}, a newly proposed implementation of KAN,  demonstrates that memory usage can be reduced from $O(LN^2G)$ to $O(LNG)$ by eliminating unnecessary input expansions, an optimization strategy that could be incorporated into MonoKAN in future work.

On the other hand, in terms of runtime, KANs tend to be slower than MLPs, despite using fewer parameters. This is largely due to the computational cost associated with evaluating and optimizing spline functions. Prior work reports training time slowdowns ranging from 6.5x to over 100x relative to MLPs \citep{Tran2024ExploringImplementation, Kundu2024KANQAS:Search}. However, recent developments have significantly narrowed this gap. In particular, the introduction of MultKAN \citep{Liu2024KANScience} has improved runtime efficiency, achieving training speeds only 3.5x slower than MLPs. Further optimizations such as GPU-compatible implementations have enabled orders-of-magnitude speedups, reducing training from hours to seconds \citep{Liu2024KANScience}. In this study, some of these advances, such as leveraging GPU computations, have been implemented. However, not all advances have been considered, as further optimizing KAN runtime is out of the scope of the paper.

While runtime efficiency remains a broader challenge for KAN-based architectures compared to MLPs, it is important to emphasize that the addition of certified monotonic constraints in MonoKAN introduces only minimal overhead relative to the standard KAN. In particular, the proposed constraints are enforced through lightweight projection operations on the network parameters as the projection is done using PyTorch clamping function \citep{Paszke2019PyTorch:Library}. This operation scales linearly with the number of parameters and has no significant impact at training time. Consequently, MonoKAN achieves certified partial monotonicity while preserving the computational efficiency of its underlying architecture.  

\begin{figure*}[h]
\centering
\includegraphics[width=\textwidth]{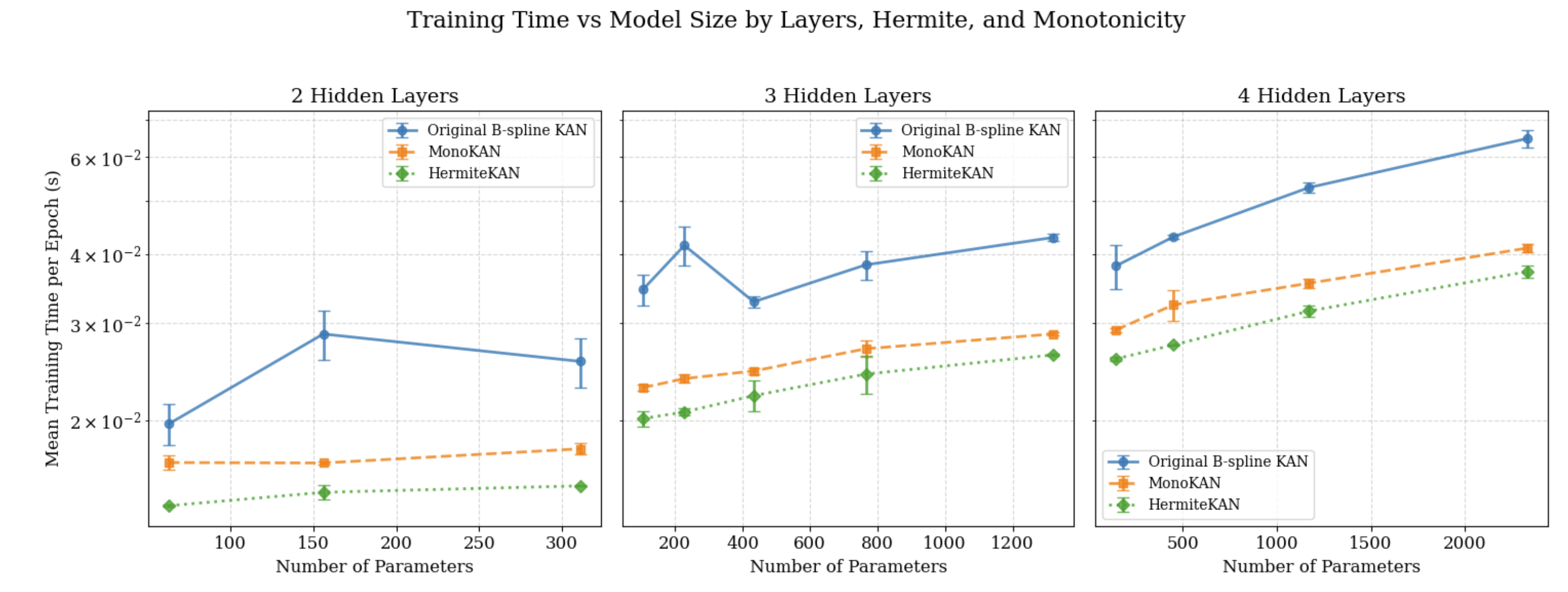}
\caption{Mean training time per epoch (in seconds) as a function of model size, measured by number of parameters, across networks with 2, 3, and 4 hidden layers. We compare the original B-spline KAN, HermiteKAN (a non-monotonic version using the proposed cubic Hermite splines), and MonoKAN (HermiteKAN with certified monotonicity constraints). Results are averaged over 5 runs with standard deviation shown as error bars. MonoKAN introduces minimal overhead relative to HermiteKAN, and both remain substantially more efficient than the original B-spline KAN across all tested configurations.}
\label{fig:training_time_vs_size}
\end{figure*}

To validate this claim, we conducted a runtime analysis comparing the training time of MonoKAN, the proposed cubic Hermite-based KAN (HermiteKAN), and the original B-spline-based KAN across a wide range of network sizes and depths. As shown in Figure~\ref{fig:training_time_vs_size}, MonoKAN introduces only a marginal increase in training time over HermiteKAN, while both remain substantially more efficient than the original B-spline formulation. For instance, even at higher model sizes with four hidden layers and over 2,000 parameters, MonoKAN's mean training time per epoch remains below $0.05$ seconds, showing a consistent runtime advantage over the B-spline KAN, which can exceed $0.06$ seconds. These results empirically confirm that the introduction of monotonic constraints has minimal impact on training efficiency when using cubic Hermite splines and that the computational overhead is negligible relative to the gains in certified monotonicity and fairness.

Additionally, considering the comparison against the baselines of certified monotonic algorithms, MonoKAN offers a practical and scalable alternative. In contrast to prior approaches such as Certified Monotonic Neural Networks \citep{Liu2020CertifiedNetworks}, which rely on Mixed Integer Linear Programming (MILP) solvers for post-hoc verification, and Counterexample-Guided Monotonic Networks \citep{Sivaraman2020Counterexample-GuidedNetworks}, which leverage Satisfiability Modulo Theories (SMT) solvers to identify violations, our method avoids solving expensive combinatorial problems. For instance, the MILP-based method proposed by \citep{Liu2020CertifiedNetworks} transforms the monotonicity verification task into an NP optimization problem where each neuron’s activation must be modeled by binary variables, leading to exponential worst-case complexity in the number of neurons and layers (e.g., $O\left(2^{\sum_{l=1}^L n_l}\right)$ for a ReLU network with $L$ layers). To mitigate intractability in deep architectures, \citep{Liu2020CertifiedNetworks} proposes to enforce monotonicity layer-wise, but this restricts the function class to compositions of monotonic functions, limiting expressiveness.

Similarly, the SMT-based approach proposed by \citep{Sivaraman2020Counterexample-GuidedNetworks} iteratively searches for monotonicity violations by formulating logical queries over network outputs. When violations are found, the corresponding counterexamples are injected into the training data with modified targets to steer the model towards monotonicity. Consequently, this process requires repeatedly querying SMT solvers and retraining the network, resulting in substantial computational overhead. Besides, both methods are also limited to piecewise linear activation functions such as ReLU, and their reliance on solver-based iterations makes them difficult to scale or extend.

In contrast, MonoKAN achieves certified monotonicity directly by design, using simple and efficient clamping operations during training, which scale linearly with the number of parameters. Furthermore, unlike MILP or SMT-based methods, which are tailored for ReLU activations and cannot be easily extended to other types of networks, MonoKAN supports a broader range of activation functions, broadening its applicability. Besides, our approach requires no additional verification steps or post-training adjustments, making it more efficient and readily deployable in practical applications.

Finally, to complement the theoretical comparison, we also performed an empirical evaluation of training time per epoch to benchmark MonoKAN against the most recent state-of-the-art monotonic models: Constrained Monotonic Networks \citep{Runje2023ConstrainedNetworks} and Expressive Monotonic Networks \citep{Nolte2022ExpressiveNetworks}. As reported in Table~\ref{tab:training_time_sec}, MonoKAN achieves superior or equal predictive performance on all three benchmark datasets while maintaining comparable training times. In particular, MonoKAN trains significantly faster than the Expressive method on the COMPAS dataset, and remains comparable on the Loan Defaulter and Blog Feedback  datasets. MonoKAN also outperforms Constrained Networks in terms of runtime in the Loan Defaulter dataset and offers better performance in all three tasks. These results support our claim that MonoKAN achieves certified monotonicity without compromising efficiency, and highlight its practical suitability for real-world applications.

\begin{table*}[htbp]
\centering
\setlength{\tabcolsep}{1pt}
\renewcommand\arraystretch{1.0}
\scriptsize{
\begin{tabular}{l||r|r|r||r|r|r||r|r|r}
\multirow{2}{*}{\bfseries Method} &
\multicolumn{3}{c||}{COMPAS}  &
\multicolumn{3}{c||}{Blog Feedback}  &
\multicolumn{3}{c}{Loan Defaulter} \\\cline{2-10}
   & Params & Acc & \shortstack{Time (s)} &
     Params & RMSE & \shortstack{Time (s)} &
     Params & Acc & \shortstack{Time (s) } \\\hline
Constrained~\citep{Runje2023ConstrainedNetworks} & 2317 & 65.3\%  $\pm$ 0.0 & 	31.80 $\pm$ 2.9 & 1101 & 0.156 $\pm$ 0.0 & 233.44 $\pm$ 15.1 & 577 & 65.0\% $\pm$ 0.0 & 163.59 $\pm$ 2.6 \\
Expressive~\citep{Nolte2022ExpressiveNetworks} & 37 & 68.7\% $\pm$ 0.0 & 101.87 $\pm$ 0.9 & 177 & 0.158 $\pm$ 0.0 & 396.34 $\pm$ 0.4 & 753 & \textbf{65.3}\% $\pm$ 0.0 & 91.7 $\pm$ 12.9 \\
MonoKAN & 4101 & \textbf{69.0\%} $\pm$ 0.0 &  48.96 $\pm$ 4.0 & 8546 & \textbf{0.155}$^1$ $\pm$ 0.0 &  526.55 $\pm$ 177.0 & 1926 & \textbf{65.3}\%$^1$ $\pm$ 0.1 &  101.87 $\pm$ 0.9\
\end{tabular}
}
\vspace{4pt}
\caption{Comparison of MonoKAN with state-of-the-art certified partial monotonic MLPs. Each “Time” cell reports the mean and the standard deviation of training time per execution (in seconds).}
\label{tab:training_time_sec}
\end{table*}

\section{Conclusion}\label{sec:conclusion} 
This paper proposes a novel artificial neural network (ANN) architecture called MonoKAN, which is based on the Kolmogorov-Arnold Network (KAN). MonoKAN is designed to certify partial monotonicity across the entire input space, not just within the domain of the training data, while enhancing interpretability. To achieve this, we replace the B-splines, proposed in the original formulation of KAN, with cubic Hermite splines, which offer well-established conditions for monotonicity and can uniformly approximate sufficiently smooth functions. Our experiments demonstrate that MonoKAN consistently outperforms existing state-of-the-art methods in terms of performance metrics. Moreover, it retains the interpretability benefits of KANs, enabling effective visualization of model behavior. This combination of interpretability and certified partial monotonicity addresses a crucial need for more trustworthy and explainable AI models.

Future research will focus on extending the architecture to splines of arbitrary degrees and investigating the effects of pruning, a key characteristic of KANs. Moreover, following the advances presented in \citep{Liu2024KANScience}, we aim to explore the integration of sparse activation mechanisms and shared spline modules into MonoKAN. These enhancements could significantly reduce memory usage and training time, making MonoKAN more scalable in high-dimensional settings without compromising its certified monotonicity guarantees. Additionally, we plan to explore whether the certified monotonicity constraints enforced by MonoKAN lead to more stable or interpretable results when evaluated with post-hoc attribution methods such as SHAP or Integrated Gradients, in order to further validate its utility in high-stakes, trust-sensitive applications. Finally, we acknowledge that in other domains, a more relaxed form of constraint (e.g., via an $\varepsilon$-tolerant monotonicity margin) may be desirable to balance interpretability with flexibility. To address such scenarios, we include as future work an extension of MonoKAN to support approximate monotonicity when full constraint enforcement is not strictly required.

\printcredits

\section*{Declaration of competing interest}
The authors declare that they have no known competing financial interests or personal relationships that could have appeared to influence the work reported in this paper.

\section*{Data availability}
All datasets used in this research are available online.
\section*{Acknowledgments}
This research was supported by funding from CDTI, with Grant Number MIG-20221006 associated with the ATMOSPHERE Project and grant PID2022-142024NB-I00 funded by MCIN/AEI/10.13039/501100011033.

\appendix
    
\section{Symbol Glossary for Theorem~\ref{teo:mono_conditions}}\label{ap:symbol_glossary}
To support the readability of Theorem~3, in this appendix we provide a detailed glossary of the mathematical symbols used throughout the monotonicity conditions. To this matter, Table \ref{tab:symbol_glossary} summarizes the key notation, including the indexing conventions for network layers, neurons, spline knots, and the associated weights and parameters. The definitions are consistent across all layers of the KAN architecture and are referenced throughout the theoretical results.


\begin{table}[h]
\small  
\setlength{\tabcolsep}{4pt}  
\centering
\begin{tabular}{l p{5cm}}  
\toprule
\textbf{Symbol} & \textbf{Meaning} \\
\midrule
$L$ & Total number of network layers \\
$l$ & Layer index, $0 \leq l \leq L-1$ \\
$I = [x^1, x^K]$ & Spline definition interval \\
$K$ & Number of spline knots in the interval of definition \\
$n_l$ & Number of neurons in the $l^{th}$ layer\\
$n_{l+1}$ &  Number of neurons in the $(l+1)^{th}$ layer\\
$i$ & Neuron index in layer $l$, $1\leq i \leq n_l $ \\
$j$ & Neuron index in layer $l+1$, $1\leq j \leq n_{l+1} $\\
$(l,i)$ & $i^{th}$ neuron in the layer $l$ \\
$(l+1,j)$ & $j^{th}$ neuron in the layer $l+1$ \\
$r$ & Index of input variable with imposed monotonicity \\
$\varphi_{l,j,i}(\cdot)$ & Spline activation connecting the $(l,i)$-neuron and the $(l+1,j)$-neuron\\
$\mathbf{b}(\cdot)$ & Base activation function (e.g. Sigmoid, Tanh, etc.) \\
$\omega^\varphi_{l,j,i}$ & Weight on spline activation connecting the $(l,i)$-neuron and the $(l+1,j)$-neuron \\
$\omega^\mathbf{b}_{l,j,i}$ & Weight on base activation function connecting the $(l,i)$-neuron and the $(l+1,j)$-neuron\\
$k$ & Spline knot index, $1 \leq k \leq K$ \\

$y^k_{l,j,i}$ & Spline value at knot $k$ \\
$d^k_{l,j,i}$ & Knot difference: $(y^{k+1}_{l,j,i} - y^k_{l,j,i})/(x^{k+1}_{l,j,i} - x^k_{l,j,i})$ \\
$m^k_{l,j,i}$ & Spline slope at knot $k$ \\
$\alpha^k_{l,j,i}$ & Norm. slope: $m^k/d^k$ \\
$\beta^k_{l,j,i}$ & Norm. slope: $m^{k+1}/d^{k+1}$ \\
\bottomrule
\end{tabular}
\caption{Glossary of symbols used in Theorem~\ref{teo:mono_conditions}.}
\label{tab:symbol_glossary}
\end{table}

\section{Proof of Theorem \ref{teo:mono_conditions}}\label{sec:ap_mono_proof}
This appendix presents a proof of Theorem \ref{teo:mono_conditions} that states a sufficient condition for a Kolmogorov Arnold Network (KAN) to be partially monotonic. First of all, let us start by presenting a proposition that states that the composition of univariate monotonic function is also monotonic.

\begin{proposition}\label{ap:prop_compo_mono}
Let $f: \mathbb{R} \to \mathbb{R}$ and $g: \mathbb{R} \to \mathbb{R}$ be two continuous functions. Then
\begin{enumerate}
    \item  $g \circ f$ is increasingly monotonic if both $f$ and $g$ are increasingly monotonic.
    \item $g \circ f$ is decreasingly monotonic if $f$ is decreasingly monotonic and $g$ is increasingly monotonic.
\end{enumerate}
\end{proposition}

\begin{proof}
\hfill

\begin{enumerate}\label{prop:ap_mono_composition}
    \item Assume $ f $ and $ g $ are increasingly monotonic. By definition, $ \forall \; x_1, x_2 \in \mathbb{R} $ such that $ x_1 \leq x_2 $, we have $ f(x_1) \leq f(x_2) $ and $ g(y_1) \leq g(y_2) $ $ \forall \; y_1, y_2 \in \mathbb{R} $ with $ y_1 \leq y_2 $. Therefore, consider any $ x_1, x_2 \in \mathbb{R} $ such that $ x_1 \leq x_2 $. Then,
    \[
    f(x_1) \leq f(x_2).
    \]
    Applying $ g $ to both sides, since $ g $ is increasing, we get
    \[
    g(f(x_1)) \leq g(f(x_2)).
    \]
    Thus, $ g \circ f $ is increasingly monotonic.
    
    \item Assume $ f $ is decreasingly monotonic and $ g $ is increasingly monotonic. By definition, $\forall \;  x_1, x_2 \in \mathbb{R} $ with $ x_1 \leq x_2 $, we have $ f(x_1) \geq f(x_2) $ and $ g(y_1) \leq g(y_2) $  $ \forall \; y_1, y_2 \in \mathbb{R} $ with $ y_1 \leq y_2 $. Consider any $ x_1, x_2 \in \mathbb{R} $ such that $ x_1 \leq x_2 $. Then,
    \[
    f(x_1) \geq f(x_2).
    \]
    Applying $ g $ to both sides, since $ g $ is increasing, we get
    \[
    g(f(x_1)) \geq g(f(x_2)).
    \]
    Thus, $ g \circ f $ is decreasingly monotonic.
\end{enumerate}

\end{proof}

Consequently,  to obtain a KAN that is increasingly (resp. decreasingly) partially monotonic with respect to the $ r^{th} $ input, it is sufficient to ensure that the $ n_1 $ activations from the $ r^{th} $ input in the first layer are increasingly (decreasingly) monotonic and that for the rest of the nodes from the following layers, where the activation function outputs generated by the $ r^{th} $ input are considered part of the input, are also increasingly monotonic. Therefore, according to the above proposition, the KAN would be increasingly (decreasingly) partially monotonic.  Considering this idea, it is obtained Theorem \ref{teo:ap_mono_conditions} that gives a sufficient condition for a KAN to be partially monotonic. 

\begin{theorem}\label{teo:ap_mono_conditions}
Given $f:\mathbb{R}^n \rightarrow \mathbb{R}$ a KAN with $L$ layers and $K$ knots in the interval of definition $I = [x^1,x^K] $, then if the basis function $\mathbf{b}$ is increasingly monotonic and the following conditions are met.
\begin{enumerate}
    \item $\omega^\varphi_{0,j,r}\geq 0, \omega^\mathbf{b}_{0,j,r}\geq 0, $ (resp. $\omega^\varphi_{0,j,r}\geq 0, \omega^\mathbf{b}_{0,j,r}\leq 0 $) \label{eq:ap_codition1}
    \item $y^{k+1}_{0,j,r}\geq y^{k}_{0,j,r}$, i.e. $d^{k}_{0,j,r}\geq 0,$ (resp. $y^{k+1}_{0,j,r}\leq y^{k}_{0,j,r}$, i.e. $d^{k}_{0,j,r}\leq 0$) \label{eq:ap_codition2}
    \item $\text{if}\;  d^{k}_{0,j,r}=0, \; m^{k}_{0,j,r} = m^{k+1}_{0,j,r} =0,\;  $\label{eq:ap_codition3}
    \item $\text{if} \;  d^{k}_{0,j,r} > 0 , \; m^{k}_{0,j,r}, \; m^{k+1}_{0,j,r}\geq 0,$ (resp. $\text{if} \;  d^{k}_{0,j,r} < 0 , \; m^{k}_{0,j,r}, \; m^{k+1}_{0,j,r}\leq 0$) \label{eq:ap_codition4}
    \item $ \text{if} \;  d^{k}_{0,j,r}  >0, \; \left(\alpha^{k}_{0,j,r}\right)^2 + \left(\beta^{k}_{0,j,r}\right)^2 \leq 9,  $\label{eq:ap_codition5}
    \item $\omega^\varphi_{l,j,i}\geq 0, \omega^\mathbf{b}_{l,j,i}\geq 0, $ \label{eq:ap_codition6}
    \item $y^{k+1}_{l,j,i}\geq y^{k}_{l,j,i}$, i.e. $d^{k}_{l,j,i}\geq 0,$\label{eq:ap_codition7}
    \item $\text{if}\;  d^{k}_{l,j,i}=0, \; m^{k}_{l,j,i} = m^{k+1}_{l,j,i} =0,\;  $\label{eq:ap_codition8}
    \item $\text{if} \;  d^{k}_{l,j,i} > 0 , \; m^{k}_{l,j,i}, \; m^{k+1}_{l,j,i}\geq 0,$\label{eq:ap_codition9}
    \item $ \text{if} \;  d^{k}_{l,j,i}  >0, \; \left(\alpha^{k}_{l,j,i}\right)^2 + \left(\beta^{k}_{l,j,i}\right)^2 \leq 9,  $\label{eq:ap_codition10}   
\end{enumerate}
where $\alpha^{k}_{l,j,i} := \frac{m^{k}_{l,j,i}}{d^{k}_{l,j,i}}$, $\beta^{k}_{l,j,i} := \frac{m^{k+1}_{l,j,i}}{d^{k+1}_{l,j,i}}$ and $1 \leq l \leq L-1, \forall \; 1 \leq k \leq K-1,1 \leq i \leq n^l,1 \leq j \leq n^{l+1}$, then 
$f$ is increasingly (resp. decreasingly) partially monotonic w.r.t the $r^{th}$ input
\end{theorem}

\begin{proof}
Let us prove the theorem by induction over the number of layers of the KAN. Without loss of generality, we will consider the case of increasing monotonicity. The case for decreasing monotonicity is followed by analogous arguments.

\textit{Base Case} $\left(n=1\right)$

Suppose that $f:\mathbb{R}^n \rightarrow \mathbb{R}$ is a KAN with 1 layer such that the KAN's structure is $[n,1]$. Therefore, by Eq. \eqref{eq:foward_pass_mod},  
\begin{equation*}
\begin{aligned}
    &\hat{y} = f(\mathbf{x}_{0}) = \sum_{i=1}^{n_0} \left( \omega^\varphi_{0,1,i} \cdot \varphi_{0,1,i}(x_{0,i}) + \omega^\mathbf{b}_{0,1,i}\cdot \mathbf{b}(x_{0,i}) \right) + \theta_{0,1}. \\
\end{aligned}
\end{equation*}
Considering conditions $\eqref{eq:ap_codition1}-\eqref{eq:ap_codition5}$ and Lemma \ref{teo:nec_condition_hermite} and Lemma \ref{teo:suf_condition_hermite}, then it is clear that $\varphi_{0,1,r}$ is monotone. Additionally, the proposed linear extrapolation of the cubic Hermite spline, with slopes $m^1$ to the left of $x^1$ and $m^K$ to the right of $x^K$, ensures that the spline $\varphi_{0,1,r}$ is \( C^1 \) continuous and monotonic across $\mathbb{R}$, not just within the data domain. Therefore, the linear combination of these monotonic functions, with positive coefficients, remains monotonic, and the composition of monotonic functions is also monotonic (by Proposition \ref{ap:prop_compo_mono}). Thus, $f$ is partially monotonic with respect to the $r^{th}$ input across $\mathbb{R}^n$.

\textit{Induction Step}. Suppose true the result for $l$ layers and let us prove it for the $(l+1)^{th}$ layer.

Considering a KAN $f:\mathbb{R}^n \rightarrow \mathbb{R}$ with structure $[n,\dots, n^l,n^{l+1}=1]$. Then by Eq. \eqref{eq:foward_pass_mod}, 
\begin{equation*}
\begin{aligned}
    &\hat{y} = f(\mathbf{x}_{0}) = \sum_{i=1}^{n_l} \left( \omega^\varphi_{l,1,i} \cdot \varphi_{l,1,i}(x_{l,i}) + \omega^\mathbf{b}_{l,1,i}\cdot \mathbf{b}(x_{l,i}) \right) + \theta_{l,1}. \\
\end{aligned}
\end{equation*}
By conditions $\eqref{eq:ap_codition6}-\eqref{eq:ap_codition10}$ and Lemma \ref{teo:nec_condition_hermite} and Lemma \ref{teo:suf_condition_hermite}, $\varphi_{l,1,i}$ is monotone $\forall \;1 \leq i \leq n^l$. Moreover, as $x_{l,i}$ is obtained from the input $\mathbf{x}_0$ as a KAN with structure $[n,\dots,n^{l-1},1]$ which satisfies all the hypothesis of the Theorem, then, by the induction hypothesis, $x_{l,i}$ is partially monotone w.r.t. the $r^{th}$ input. Therefore, considering Proposition \ref{ap:prop_compo_mono} and the same reasoning as in the base case, $f$ is partially monotone w.r.t. the $r^{th}$ input across $\mathbb{R}^n$.
\end{proof}

\section{Dataset Descriptions}\label{ap:data_description}

This appendix provides a detailed overview of the datasets used in our experiments. For each dataset, we specify the task type, input dimensionality, number of monotonic features, and the size of the training and test sets.

\begin{table*}[h]
\centering
\caption{Summary of datasets used in the experiments}
\begin{tabular}{lcccc}
\toprule
\textbf{Dataset} & \textbf{Task} & \textbf{Feature Dim.} & \textbf{Mono Features} & \textbf{Train/Validation/Test Size} \\
\midrule
COMPAS & Classification & 13 & 4 & 3949 / 988 /  1235 \\
Blog Feedback & Regression & 276 & 8 & 37841 / 9461 / 6968 \\
Loan Defaulter & Classification & 28 & 5 & 334957 / 83740 / 70212 \\
Auto MPG & Regression & 8 & 3 & 250 / 79 / 80 \\
Heart Disease & Classification & 13 & 2 & 193 / 49 / 61 \\
\bottomrule
\end{tabular}
\label{tab:dataset_summary}
\end{table*}

\textbf{COMPAS:} The COMPAS dataset \citep{Angwin2016MachineBlacks.} contains criminal history and demographic information for 6,172 individuals arrested in Florida. The objective is to predict the likelihood of recidivism within two years, based on 13 input features, which is modeled as a binary classification problem. To reflect ethical considerations, monotonicity is enforced with respect to four features that indicate criminal history: number of prior adult convictions, number of juvenile felonies, number of juvenile misdemeanors, and number of other convictions. Higher values for these variables should correspond to an increased predicted risk.

\textbf{Blog Feedback:}  
The Blog Feedback dataset \citep{Buza2014FeedbackBlogs} includes over 54,000 examples derived from blog posts, with the goal of predicting the number of comments a post will receive in the 24 hours following publication. Each sample consists of 276 features extracted from the post metadata and publication context. Eight of these features (A51, A52, A53, A54, A56, A57, A58, and A59) are expected to be monotonically non-decreasing with the target variable. Following common practice, we filter out extreme outliers by removing samples with targets above the 90th percentile to improve robustness to skewed error metrics such as MSE.

\textbf{Loan Defaulter:}\footnote{https://www.kaggle.com/datasets/wordsforthewise/lending-club/code}  
Based on publicly available lending data covering loans issued between 2007 and 2015, this dataset includes 488,909 entries with 28 input features. These features span credit history, income data, and loan status details. The task is to predict whether a borrower will default. Monotonicity constraints are imposed on a subset of features with strong financial interpretation: the model should predict increasing default probability with more public bankruptcies and higher debt-to-income ratio, and decreasing default probability with higher credit scores, longer employment history, and greater annual income.

\textbf{Auto MPG:}\footnote{https://archive.ics.uci.edu/dataset/9/auto+mpg}  
This dataset includes fuel efficiency data for various car models. The prediction task is to estimate miles-per-gallon (MPG), framed as a regression problem. The input consists of 8 features, including engine and vehicle specifications. Monotonicity is enforced for three attributes: weight, horsepower, and displacement, all of which are assumed to be negatively correlated with MPG due to mechanical and energy-efficiency constraints. The dataset is sourced from the UCI Machine Learning Repository.

\textbf{Heart Disease:}   \footnote{https://archive.ics.uci.edu/dataset/45/heart+disease}
The Heart Disease dataset contains 303 patient records with 13 medical features. The task is to classify the presence or absence of heart disease. Based on medical rationale, monotonicity constraints are applied to two input variables: trestbps and cholesterol. Higher values for these features are assumed to increase the risk of heart disease, and the model is designed to reflect this domain knowledge

\bibliographystyle{cas-model2-names}
\bibliography{references}



\end{document}